\documentclass[sigconf,nonacm]{acmart}
\setcopyright{none}
\usepackage{amsmath}
\usepackage{graphicx}
\usepackage{xcolor}
\usepackage{booktabs}
\usepackage{makecell}
\usepackage{braket}
\usepackage{tikz}
\usetikzlibrary{decorations.pathreplacing, calc, positioning}

\usepackage[dvipsnames]{xcolor}
\newcommand{\accepted}[1]{{\color{black} #1}}

\title{A Quantum Machine Learning Approach to   Content-Based Media Recommendation}

\author{Mina Abbaszadeh}
\authornote{These authors contributed equally to this work.}
\authornote{Corresponding authors: Mina Abbaszadeh
(m.abbaszadeh@ucl.ac.uk) and Mehrnoosh Sadrzadeh
(m.sadrzadeh@ucl.ac.uk).}
\affiliation{
  \institution{University College London}
  \country{United Kingdom}
}

\author{Ouissal Moumou}
\authornotemark[1]
\affiliation{
  \institution{Quantinuum}
  \country{United Kingdom}
}

\author{Anna Pearson}
\affiliation{
  \institution{Quantinuum}
  \country{United Kingdom}
}

\author{Bob Coecke}
\authornote{This work was carried out whilst at Quantinuum}
\affiliation{
  \institution{Relational Intelligence}
  \country{United Kingdom}
}

\author{Rob Cooper}
\affiliation{
  \institution{British Broadcasting Corporation}
  \country{United Kingdom}
}

\author{Alessandro Piscopo}
\affiliation{
  \institution{British Broadcasting Corporation}
  \country{United Kingdom}
}

\author{\mbox{Mehrnoosh Sadrzadeh}}
\authornotemark[2]
\affiliation{
  \institution{University College London}
  \country{United Kingdom}
}

\begin{document}

\begin{abstract}
We use quantum machine learning techniques to develop a  content-based media recommender system and evaluate its performance on TV programmes using textual descriptions from one week of BBC iPlayer data. Grounded in the Quantum Natural Language Processing framework (QNLP), programme synopses are mapped to variational quantum circuits using their semantic structure. The parameters of these circuits are trained via a supervised pairwise similarity learning task.

We benchmark the quantum recommender against a classical BiLSTM baseline trained on the same dataset and objective. While both models achieve comparable performance on the similarity learning task, the quantum model requires significantly fewer parameters (47k vs. 17.8M). 
When the BiLSTM is constrained to a matching parameter budget, its performance degrades substantially (by 3.63 percentage points in Mean Average Precision (MAP)), indicating that the structured quantum representation provides a highly parameter-efficient alternative in this setting. 
We demonstrate the feasibility of executing the trained models on quantum hardware by deploying the circuits on Quantinuum’s System Model H2 (and accompanying emulator) using a SWAP-test-based protocol. 
The results confirm that the similarity signal is preserved under realistic, noisy conditions. Taken together, these findings highlight the potential of variational quantum circuits for recommendation tasks and represent a practical step toward their real-world deployment.
\end{abstract}

\maketitle

\section{Introduction}
In the rapidly evolving landscape of Video-on-Demand (VoD), defined by the recent UK Media Act 2024 \cite{medAct}, public service broadcasters like the BBC face a dual challenge: ensuring content prominence while delivering hyper-personalized user experiences (Ofcom, 2026 \cite{Ofcom}). Content-based recommender systems are central to this mission, particularly in addressing the `cold-start' problem, where new content lacks the interaction history required for collaborative filtering  \cite{AdomaviciusTuzhilin2005,Weietal2017,Zawiaetal2025}.

However, modern content-based recommenders  struggle with the explosive growth of content metadata. Classical deep learning models often encounter the `readout problem' and high-dimensional feature sparsity when mapping thousands of niche attributes across millions of items \cite{Sim2019Expressibility}. This leads to over-specialization and a lack of `serendipity' in user discovery.

Quantum Machine Learning (QML) offers a paradigm shift. While early foundational work focused on polylogarithmic run times for matrix factorisation and explored their applications to collaborative filtering recommender systems \cite{KerenidisPrakash2017}, this paper  identifies a more immediate benefit via using quantum features and proposes a quantum content-based recommender. 

The proposed methodology leverages \emph{Quantum Natural Language Processing} (QNLP)\cite{coecke2020foundations,zeng2016quantum,lorenz2023qnlp}  to transform program synopses   into quantum states using their semantic content. Unlike standard NLP, which represents semantics in high-dimensional Euclidean space, this approach encodes textual assets directly onto a variational quantum circuit (VQC) and represents semantics in Hilbert space. The work uses the QNLP   software  \emph{Lambeq}~\cite{krawchuk2025efficient,Liu2023Pipeline,kartsaklis2021lambeq,Duneau2024QDisCoCirc,coecke2021discocirc}, which extracts the  semantic relations within a piece of text and  embeds them  into the gate structure of a VQC.  The parameters of the circuits are learnt by maximising the overlap between the quantum circuits,
leading to an objective function based on the  semantic similarity between program synopses. The similarities are subsequently inputted to a standard KNN recommender system to generate recommendations and are evaluated on a BBC iPlayer benchmark of a week's TV programmes. A subset of the test set is also evaluated on a quantum computer: Quantinuum's H2 device~\cite{MosesH2}.

We trained two QNLP models: an ideal formulation (evaluated with post-selection) and a practical  variant (evaluated with discard).  With only 47K parameters, both quantum models significantly outperformed a parameter-matched BiLSTM baseline (13.33\%  \accepted{for the ideal model and 12.93\% for the practical one} vs. 9.40\% MAP). The ideal QNLP model also marginally exceeded the performance of the full classical BiLSTM equipped with 17.8M parameters (13.33\% vs. 13.03\% MAP). While the practical model experienced a minor performance reduction to 12.93\% MAP, it remained substantially superior to the parameter-matched classical baseline.  During training, both quantum variants demonstrated stable convergence, achieving training, validation, and test accuracies in the high 80\% and  90\% range. In contrast, the parameter-matched BiLSTM struggled to learn effectively, with its train, validation, and test metrics lingering near chance level. The full-scale BiLSTM converged successfully, reaching classification accuracies comparable to those of the quantum models. To assess the importance of semantic structure, we also benchmarked against a Quantum Bag-of-Words baseline (\textbf{Q-BOW}), which encodes unordered sets of words without accounting for semantic sentence or text structure;  this also  performed just above chance level at 58.58\% and failed to learn effectively. 

The performance of the quantum models improved substantially when integrated with classical content-based baselines, specifically neural topic representations. This hybrid ensemble yielded a 0.46 point gain in MAP, reaching from 12.93\% to 13.39\% and narrowing the gap with collaborative filtering (16.00\%). In contrast, combining the classical BiLSTM with these same content-based models provided a smaller performance benefit, improving from 13.03\% to 13.44\% MAP. Complementary evaluation metrics, including NDCG, Intra-List Diversity (ILD), surprisal, and catalog coverage, exhibited trends consistent with the MAP findings. \textbf{Q-BOW} exhibited poor generalization: although it attained a training accuracy of 71.79\%, its validation and test accuracies degraded to 58.58\% and 65.65\%, respectively. Furthermore, its downstream recommendation performance was poor, yielding a MAP of just 9.30\%---falling below even the parameter-matched BiLSTM baseline. It is worth noting that despite the fact that the classical content-based baseline was trained on the full transcripts of the programmes rather than short synopses, its MAP was only 0.3\% above the practical quantum model and lagged behind the ideal quantum model by 0.1\%.

When executed on the Quantinuum H2 platform, both noisy-emulator and physical-hardware executions preserved the predicted similarity structure. We validated this using the Pearson correlation coefficient ($r$). The noisy-emulator SWAP-test overlaps remained strongly correlated with the classical training fidelity ($r = 0.875$), compared with $r = 0.910$ for the noiseless emulator. On the H2-2 hardware subset, the measured overlaps correlated strongly with the training fidelity ($r = 0.963$) and with the corresponding noisy-emulator results ($r = 0.952$). These results demonstrate the viability of variational quantum circuits for real-world tasks such as recommendation generation. Importantly, they show that quantum models can achieve competitive performance with orders of magnitude fewer parameters than classical deep learning architectures such as BiLSTMs, using thousands rather than millions of parameters. This parameter efficiency and resilience to physical device noise represent a promising step toward deploying quantum algorithms in practical commercial settings.

The remainder of this paper is structured as follows. We begin by reviewing the foundational framework of QNLP, detailing how programme synopses are translated into VQCs. Next, we describe the computation of content-based similarities between these synopses and their use in training the VQCs. We then outline our experimental setup and implementation details, followed by the recommendation generation results. Finally, we present a hardware feasibility study on Quantinuum’s H2 device, comparing the correlation between classical simulations, the quantum emulator, and real hardware execution.

\section{Background: Overview of QNLP}

In this section, we review the theoretical foundations underlying the QNLP framework and the process of converting text into VQCs. We use two frameworks with QNLP; these are called  \emph{DisCoCirc} and {QDisCoCirc}. DisCoCirc  analyses the semantic content of  text and represents these by diagrams that facilitate the transition to quantum circuits~
\cite{Liu2023Pipeline,kartsaklis2021lambeq}. In this framework, sentences are diagram fragments that are  composed to form novel discourse-level diagrams. QDisCoCirc extends this framework to the quantum setting by mapping the diagrams into Variational Quantum Circuits (VQCs) via a structure-preserving mapping \cite{Duneau2024QDisCoCirc}.

\paragraph{From text to diagrams.}
Within each sentence, entities correspond to wires and the relations between them to boxes; these are transformations of entities. Discourse-level text diagrams are obtained by composing sentence diagrams sequentially.
Here, these entities are represented as dynamic objects, allowing sentence diagrams to be composed with one another.

In controlled or task specific settings, diagrams  are further decomposed to a set of  \emph{generators} that represent the words and phrases within each sentence.

The main generators are as follows:
\begin{itemize}
    \item \textbf{States}, representing entities (e.g., noun phrases),
    \item \textbf{Boxes}, representing predicates and relations,
    \item \textbf{Effects}, representing semantic tests or queries,
    \item \textbf{Discard operations}, representing the removal of irrelevant subsystems.
\end{itemize}

\begin{figure}[!h]
    \centering
    \includegraphics[
    width=0.9\linewidth,
    height=0.9\textheight,keepaspectratio]
    {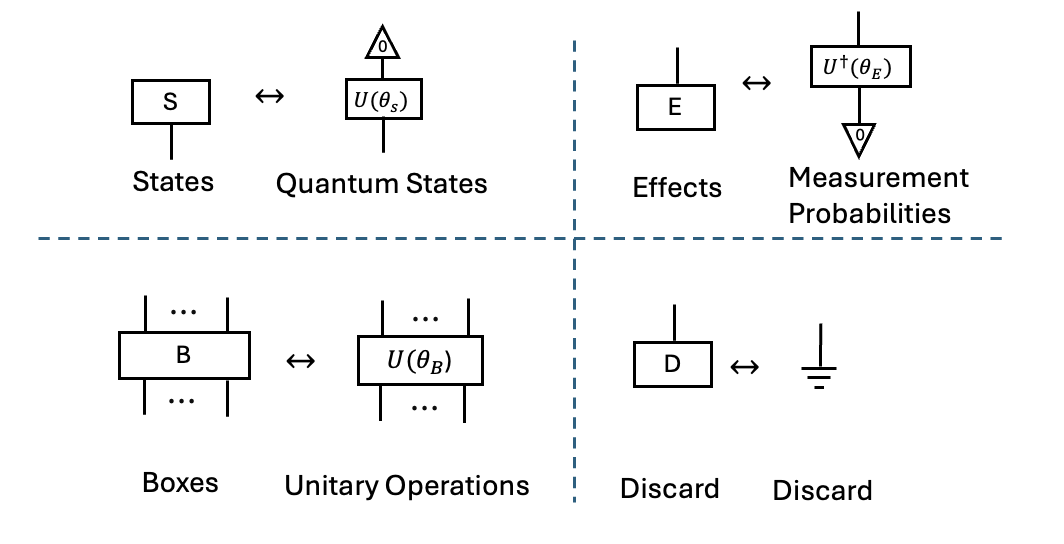}
    \caption{Basic generators in the DisCoCirc framework and their quantum realisation.}
    \label{generators}
\end{figure}

The generators are combined by sequential and parallel composition, depending on the flow of information within text; if the information is updated, sequential composition is used, while if it is untouched and some other independent piece of information is provided, parallel composition is used. Sequential composition connects the outputs of one diagram to the inputs of the next, while parallel composition places components side by side without connecting their inputs and outputs. The above generators are depicted on the left hand sides of the four dashed regions of Figure \ref{generators}.

\begin{figure*}[h!]
    \centering
    \includegraphics[
    width=0.85\linewidth,
    height=0.85\textheight,keepaspectratio]
    {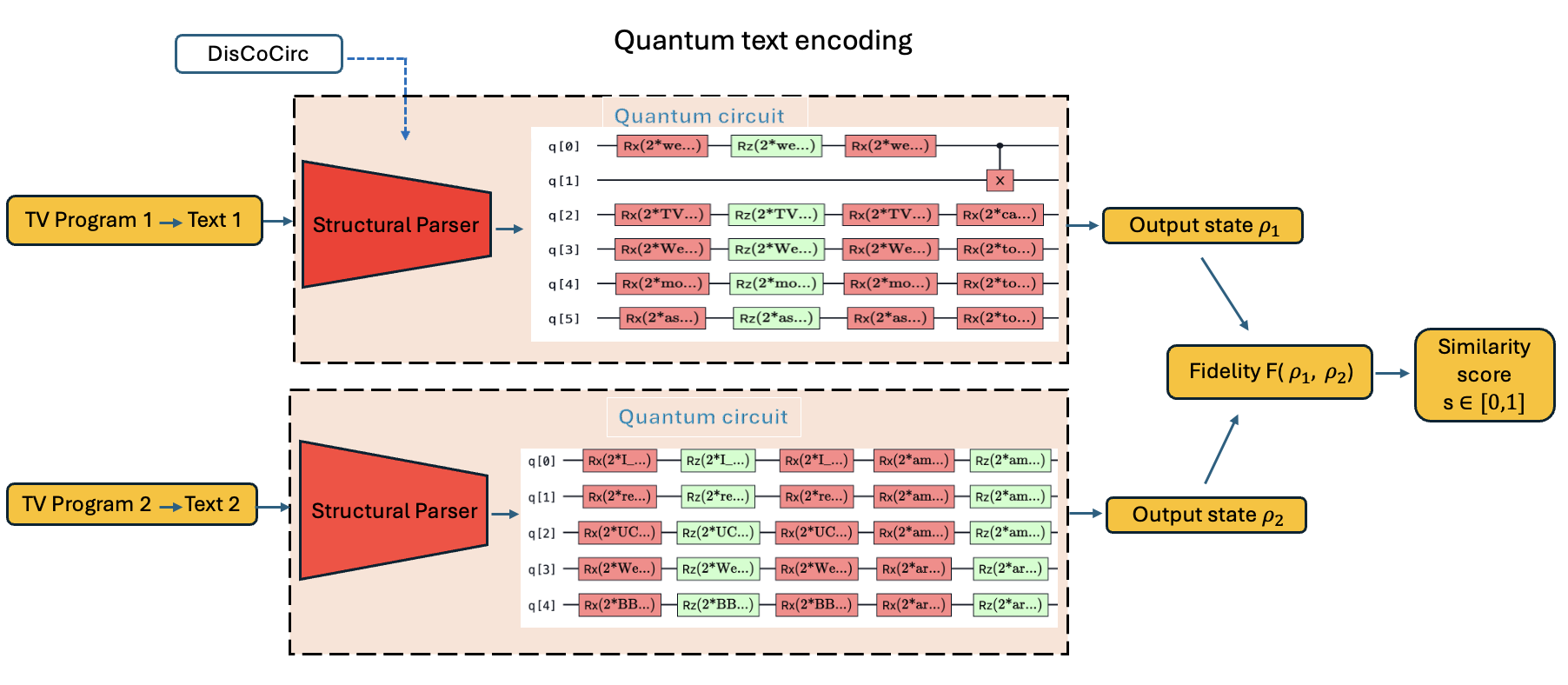}
    \caption{Pipeline for quantum text-similarity computation. Each TV programme synopsis is structurally parsed and mapped to a parameterised quantum circuit, producing output states $\rho_1$ and $\rho_2$. Their fidelity $F(\rho_1,\rho_2)$ yields the final similarity score $s \in [0,1]$.}
    \label{pipeline}
\end{figure*}

\paragraph{From Diagrams to VQCs.}
Text diagrams are mapped to  VQCs via the  mapping depicted  in Figure \ref{generators}. The structure of each resulting circuit preserves the structure of its corresponding diagram, which in turn represents the semantic relations within and between the sentences of a piece of text.  In this mapping:
\begin{itemize}
    \item each wire is assigned a fixed number of qubits,
    \item states correspond to parameterised quantum state preparations,
    \item boxes correspond to parameterised quantum operations (in the most general case the boxes can have ancillas),
    \item effects correspond to measurements of specific outcomes, or equivalently to state unpreparations,
    \item discard operations correspond to removing subsystems.
\end{itemize}

Formally speaking, each circuit represents a series of tensors and tensor contractions, in such a way that the circuit as a whole (both before and after the contractions) corresponds to a state in a Hilbert space~\cite{Stoudenmire2016,Biamonte2017}.  Circuit parameters are  optimised via classical simulation of these states, which are subsequently  executed on quantum hardware for inference.

\section{Methodology}\label{Methods}
In this section, we go over our methodology for training the  VQCs on the programme synopses of the BBC iPlayer, built according to the QNLP approach.

\subsection{Task Definition and Dataset}
We use a  dataset consisting of a week of TV programmes on iPlayer, where each programme is associated with a medium-length textual synopsis.  These are 2-3 sentences  of 5-10 words each that  provide a  summary of the content of each TV programme. The dataset also consists of  users' behavioural data, also observed within a given week. This data shows which user watched which programme for more than 5 minutes.  The textual data is used to construct variational quantum circuits for the content-based quantum recommender. The behavioural data was used by the BBC Research and Development recommender software to provide gold standard collaborative filtering recommendations for comparison with the quantum model. This data is highly confidential and is not provided to third parties.  

We implement a pairwise, content-based similarity learning task over programme synopses. The objective is to learn a mapping from a pair of synopses to a similarity score $\in [0, 1]$, representing their semantic relatedness. Each synopsis is encoded into a VQC following the procedure in the previous section. The semantic similarity between two synopses is then computed via the quantum state overlap of their respective VQCs (formalized below). Figure~\ref{pipeline} illustrates this pairwise similarity computation, in which each programme synopsis is structurally parsed and mapped to a parameterised quantum circuit, and the similarity between the resulting quantum representations is computed using fidelity. Finally, the VQC parameters are trained on ground-truth \accepted{binary} similarity labels generated by a neural topic model (NTM-R). We select this method because it leverages both latent features within the synopses and neural semantic embeddings of the words associated with each topic. The NTM-R model is trained on the full transcript of each programme—a multi-page, document-length text that yields richer similarity estimates than short synopses alone.

\begin{table*}[h!]
    \centering
    \renewcommand{\arraystretch}{1.3}
    \setlength{\tabcolsep}{6pt}
    \begin{tabular}{l l  c c c c c}
        \toprule
        \textbf{Models} &  \textbf{Train} & \textbf{Valid} & \textbf{Test} & \textbf{mean train} & \textbf{mean test} & Trainable Parameters\\
        \midrule
        \textbf{QNLP: discard} & 98.34\% & 89.00\% & 90.90\% & 98.43\% $\pm$ 0.49\% & 90.66\% $\pm$ 2.93\% & 47.4K\\
        \midrule
        \textbf{QNLP: post-selection} & 96.69\% & 87.87\% & 88.88\% & 97.44\% $\pm$ 0.7\% & 91.47\% $\pm$ 1.77\% & 47.4K\\
        \midrule
          \textbf{Q-BOW} & 71.79\% & 58.58\% & 65.65\% & 72.59\% $\pm$ 6.34\% & 67.11\% $\pm$ 10.90\% & 46.4k \\
        \midrule
        \textbf{BiLSTM@17M} &  92.76\% & 88.89\% & 90.91\%& 95.59\% ± 0.66\% & 91.70\% ± 1.87\% & 17.8M\\
        \midrule
        \textbf{BiLSTM@47K} & 53.24\% & 52.52\% & 53.54\% & 53.30\% $\pm$ 0.86\% & 53.30\% $\pm$ 3.09\% & 47.8K\\
        \bottomrule
    \end{tabular}
    \caption{Results for the pairwise similarity prediction task. Train/Valid/Test results correspond to the initial fixed-split experiment used for hyperparameter tuning. Mean train and mean test report 5-fold cross-validation performance (mean ± standard deviation) using the selected hyperparameters. QNLP is our main quantum recommender; Q-BOW is a quantum baseline that only considers word meaning; the @ signs of the BiLSTMs are their numbers of parameters. For reference, a majority-class baseline that always predicts the positive class achieves an accuracy of 53.3\%}
    \label{tab:pairwise_similarity}
\end{table*}

\subsection{NTM-R Similarities and Label Construction}

NTM-R stands for Neural Topic Model with Regularization~\cite{NTMR}; it is a deep learning approach that identifies hidden topics in text. It works by encoding each document into a set of topic proportions and decoding these back into words. A regularization component based on pre-trained word embeddings helps the model form clearer and more coherent topics.

We use the NTM-R implementation of previous work~\cite{CagaliWazniNazirSadrzadehNewell2023,CagaliSadrzadehNewell2021}, where each TV program is represented using a bag-of-words (BoW) vector created from its subtitle text. The subtitles are tokenized, cleaned, and converted into a vocabulary of all unique words appearing across the dataset. For each program, the BoW vector records how often each word occurred, producing a numerical representation of its textual content. These vectors are  inputted into the NTM-R algorithm, which in turn learns abstract topics and represents each document as a mixture of those topics. The algorithm follows the structure of a variational autoencoder, consisting of two main parts: an encoder and a decoder, described below:

\begin{itemize}
    \item 
    The encoder takes the BoW vector of a program and compresses it into a lower-dimensional latent vector that captures how strongly the program relates to each topic.
    \item 
    The decoder reconstructs the original word distribution from this latent vector and learns which words are most associated with each topic.
\end{itemize}

An additional topic coherence regularization term was added to the autoencoder to guide the model to form more meaningful and interpretable topics. This was done  by using pre-trained word embeddings (\cite{CagaliSadrzadehNewell2021} used the statistical distributional vectors known as GloVe \cite{Glove}) to ensure that the top words within each topic are semantically similar. Then, each TV program was represented by the latent topic distribution vector, which is a numerical summary indicating what proportion of each topic was presented in that program. 

\subsection{Circuit Construction and Ansatz}

There exist several VQCs ans\"atze proposed, e.g. see~\cite{IQP,HEA,Sim2019Expressibility}. In this work, we adopt the Sim4 ansatz \cite{Sim2019Expressibility}, which provides a structured parameterisation of single-qubit rotations and entangling operations.  This choice allows us to control circuit depth and the number of trainable parameters, while retaining the level of expressivity required for our model. The number of layers of the ansatz is a hyperparameter of the model and can be increased by stacking multiple layers of parameterised gates. 

States are implemented using  a standard Euler angle parametrisation, where a qubit is parametrised by a sequence of Z-X-Z rotation gates. Effects are measurement probabilities; boxes are unitary operations parametrised by the ansatz. Discard operations remove the corresponding subsystem by taking the partial trace over its qubits, leaving a reduced density matrix on the remaining output qubits.

To evaluate the similarity of a pair of quantum circuits, we project the \( 2^n \)-dimensional output vector space of each circuit into 4-dimensional vector space by merging the output wires with a box. For each circuit we retain two output qubits; the remaining qubits are  discarded. \accepted{Thus, the merge box acts as a global aggregation layer over the (latent) qubits present in the text circuit to extract a fixed-length embedding.}

\subsection{Training Objective and Optimisation}

We implement the quantum recommender using the \texttt{PytorchQuantum} package from the lambeq library, which relies on PyTorch as the backend for all tensor operations, and we use the \texttt{Adam} optimiser.  The objective function is implemented via the margin-based contrastive loss formula given below:

\[
L = 0.5 \left( y (1 - s)^2 + (1 - y) \max(0, s - \text{margin})^2 \right).
\]

\noindent
During training, a continuous similarity score $s \in [0,1]$ is compared with a ground-truth label $y \in \{0,1\}$, which provides the supervision signal. The margin is set to $0.5$. The continuous similarity score is the overlap between two quantum states and the ground-truth label comes from the NTM-R similarities. \accepted{The NTM-R similarities are converted into binary labels by ranking the similarities for each programme. The top 20 most similar programmes are labelled 1, while the bottom 20 least similar programmes are labelled 0.} \accepted{The programme pairs in between the top 20 and bottom 20 do not feature in the dataset.} 
\accepted{By training on labels derived from the most and least similar NTM-R pairs, we aim to learn a similarity measure that is correlated with NTM-R on these extreme examples but is ultimately distinct, since the quantum model learns its representation directly from the programme synopses. Intermediate programme pairs are not used for supervision, allowing the learned model to generalise beyond the NTM-R derived training subset.}

\paragraph{The Uhlmann Fidelity Metric over Density Matrices}
The overlap between two quantum states is computed via \emph{Uhlmann} fidelity.  Each circuit produces a density matrix representation $\rho \in \mathbb{C}^{d \times d}$ obtained via tensor network contraction. The Uhlmann fidelity between two quantum states $\rho_a$ and $\rho_b$ is defined as follows:
\[
F(\rho_a, \rho_b) = \left( \text{Tr} \sqrt{\rho_a^{1/2} \rho_b \rho_a^{1/2}} \right)^2.
\]
The matrix square roots are computed via Eigendecomposition to ensure the metric remains differentiable during the parameter optimization process. The resulting fidelity $F \in [0, 1]$ is what is used as the similarity in our contrastive loss formula above, denoted by  $s$. The Uhlmann fidelity generalizes the pure-state overlap to mixed-state representations, enabling the model to learn optimal circuit parameters by maximizing the density matrix overlap of two semantically related circuits.

To design a recommender system viable on near-term quantum hardware, we must account for mixed states alongside pure states. In QNLP,  post-selection or discarding are the method used to obtain a single output. While post-selection yields a pure state vector that is straightforward to model, its exponentially low success probability makes it prohibitively expensive on any quantum device.  Conversely, discarding qubits traces out entangled subsystems, transforming the global pure state into a local mixed state represented by a density matrix. Consequently, we train two distinct circuit models: an idealized post-selected baseline and a hardware efficient partial trace model.

\begin{table*}[h!]
    \centering
    \renewcommand{\arraystretch}{1.3}
    \setlength{\tabcolsep}{6pt}
    \begin{tabular}{l c c c c c}
        \toprule
        \textbf{Content-Based Models} & \textbf{MAP\%} & \textbf{NDCG\%} & \textbf{ILD\%} & \textbf{Surprisal} & \textbf{Coverage\%} \\
        \midrule
        \midrule
        Quantum Models:\\
        \textbf{QNLP: discard} & 12.93 & 25.87 & 78.84 & 5.31 & 100.00 \\
        \textbf{QNLP: post-selection} & 13.33 & 26.80 & 78.28 & 5.30 & 100.00 \\
        \textbf{Q-BOW} & 9.30 & 20.58 & 79.31 & 5.63 & 100.00 \\
        \midrule
        Classical Models:\\
           \textbf{BiLSTM@17.8M} & 13.03 & 26.02 & 79.10 & 5.37 & 100.00 \\
              \textbf{BiLSTM@47K} & 9.40 & 20.61 & 79.88 & 5.84 & 100.00 \\
            \textbf{NTM-R} & 13.23 & 26.53 & 78.88 & 5.24 & 100.00 \\
               \midrule
       Hybrid Models  \\
       \midrule
       One Quantum + One Classical Model:\\
    \textbf{QNLP:discard + NTM-R} & 13.39 & 26.78 & 78.66 & 5.20 & 100.00 \\
    \textbf{QNLP:post-selection + NTM-R} & 13.80 & 27.47 & 78.08 & 5.19 & 100.00 \\
    \textbf{QNLP:discard +BiLSTM} & 13.32 & 26.46 & 78.89 & 5.32 & 100.00 \\
    \textbf{QNLP:post-selection +BiLSTM} & 13.43 & 26.80 & 78.44 & 5.32 & 100.00 \\
    \midrule
     One Quantum + Two Classical Models: \\
    \textbf{QNLP:discard +BiLSTM + NTM-R} & 13.35 & 26.70 & 78.68 & 5.24 & 100.00 \\
    \textbf{QNLP:post-selection +BiLSTM + NTM-R} & 13.73 & 27.26 & 78.26 & 5.25 & 100.00 \\
  \midrule
  Two Classical Models:
            \textbf{BiLSTM + NTM-R} & 13.44 & 26.84 & 78.56 & 5.23 & 100.00 \\
        \midrule
        \midrule
        \textbf{Collaborative Filtering Model} & 16.00 & 30.74 & 79.16 & 4.10 & 98.58 \\     
        \bottomrule
    \end{tabular}
    \caption{Evaluation of accuracy and diversity metrics for 145 TV programs.}
    \label{tab:not-hybrid}
\end{table*}

\section{Evaluation and  Results on Pairwise Similarity}

To obtain a robust estimate of generalisation performance, we employ 5-fold cross-validation over the complete set of 985 selected programme pairs. The dataset consists of 985 pairs, which are partitioned into 5 disjoint folds of approximately equal size ($\approx$197 samples per fold). In each iteration, one fold is used as the held-out test set, while the remaining four folds ($\approx$790 samples) are used for training. The model is re-initialised from scratch for each fold to ensure independence across runs and to prevent information leakage between partitions. This procedure ensures that each sample is used exactly once for testing and four times for training.  For evaluation, similarity scores are thresholded to obtain binary predictions. The threshold was selected during preliminary experiments on a fixed train/validation split and is kept fixed during the 5-fold cross-validation \accepted{at 0.5}. Binary predictions are then compared with the ground-truth labels to compute accuracy.
\accepted{The five randomly generated folds were approximately balanced with respect to the ground-truth labels, with each test fold containing comparable numbers of positive and negative pairs. }
\accepted{The dataset contains 53.3\% positive and 46.7\% negative pairs. Therefore, the appropriate constant-prediction baseline is 53.3\%: a model that predicts the same label for every pair achieves its highest accuracy by always predicting the majority positive class.} 
\accepted{The present cross-validation therefore evaluates generalisation to unseen programme pairs rather than to completely unseen synopses. Given the relatively small size of the dataset, we did not construct a separate evaluation setting for unseen synopses. Generalisation to unseen synopses composed of previously observed vocabulary will be considered in future work.}

\begin{figure*}[h!]
    \centering
    \includegraphics[
    width=0.9\linewidth,
    height=0.9\textheight,keepaspectratio]
    {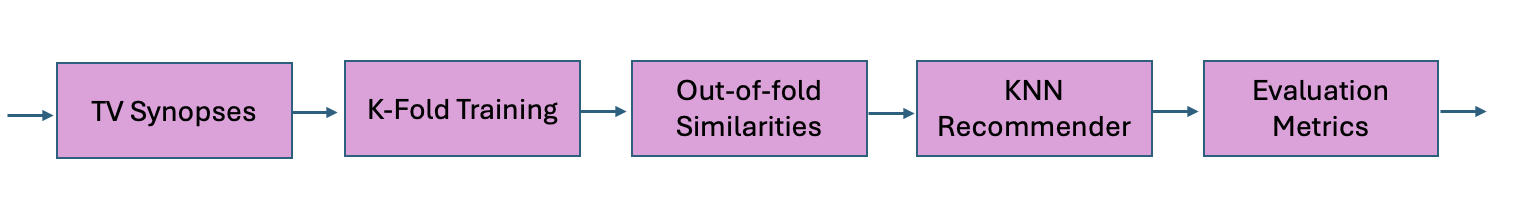}
    \caption{Pipeline for evaluating learned similarity within a content-based recommender.} 
    \label{quantum-pipeline-recommender}
\end{figure*}

Each fold is trained for 25 epochs using fixed hyperparameters determined in preliminary experiments. Final performance is reported as the mean and standard deviation of test accuracy across the five folds. Figure~\ref{quantum-pipeline-recommender} illustrates the overall training pipeline. The first four rectangles on the left describe the training process, the last two the recommendation generation procedure;  we will explain the latter later. 

Hyperparameters (learning rate, seed, batch size, margin parameter, and number of epochs) were selected using preliminary experiments on a fixed train/validation/test split of 787, 99, and 99 programme pairs, respectively. The models were trained for up to 50 epochs to analyse convergence behaviour. We observed that training consistently converges within 20–25 epochs. Based on this observation, we fix the number of training epochs to 25 for cross-validation experiments, reducing computational cost while maintaining performance. No further hyperparameter tuning was performed during cross-validation.

We use a batch size of 16 and a learning rate of $3.45 \times 10^{-4}$. While the experiments do not require specialised hardware, training time is substantial, thus all experiments were conducted on UCL CS's  high-performance computing  cluster using CPU-based execution, with each run allocated 16\,GB of RAM and a single CPU core.  

A full 5-fold cross-validation run required approximately 70–80 hours of wall-clock time and the maximum allocated runtime was 96 hours.

To compare the performance of the quantum recommender with a classical baseline, we implement a bidirectional Long Short-Term Memory (BiLSTM) neural network on the same task. 

The BiLSTM was trained  from scratch, with no pre-trained embeddings or external knowledge to ensure a fair comparison with the quantum model, which likewise learns its representations exclusively from the training data.
Each synopsis is tokenised by white space and mapped to a learned embedding of dimension $d_e = 64$.  Sequences are padded to a uniform length with a \texttt{<PAD>} token. The embedded sequence is then processed by two-layers  with hidden dimension $d_h = 128$. This BiLSTM is described below:

\begin{equation}
    \overrightarrow{h}_t, \overleftarrow{h}_t = \text{BiLSTM}(w_t,\; h_{t-1}), \qquad \mathbf{s} = [\overrightarrow{h}_T \,\|\, \overleftarrow{h}_1],
\end{equation}
where $w_t \in \mathbb{R}^{d_e}$ is the token representation and $\mathbf{s} \in \mathbb{R}^{2d_h}$ is the  sentence representation. The degree of similarity between two sentence representations is computed using cosine similarity, and the model is trained using the same margin-based contrastive loss described in Section~\ref{Methods}. For optimisation we used Adam again, with a learning rate of $10^{-3}$ and batch size 16. Dropout is applied at rate 0.3 for regularisation.  Training ran for up to 100 epochs with early stopping on validation loss. \accepted{For the parameter-matched BiLSTM, our objective was to match the overall number of trainable parameters of the QNLP model rather than the dimensionality of the final synopsis representation. We therefore used a token embedding dimension $d_e=18$, a hidden dimension $d_h=36$, one bidirectional LSTM layer, and a final projection dimension $d_p=1$. For this architecture, the number of trainable parameters is $ P = Vd_e + 8d_h(d_e+d_h+2) + 4d_hd_p + 2d_p^2 + 7d_p, $ where $V$ denotes the vocabulary size. For $V=1754$, this gives $47{,}853$ trainable parameters, approximately matching the $47.4$K parameters of the QNLP model.}

The results are summarized in Table~\ref{tab:pairwise_similarity}. Here, \textbf{QNLP: \allowbreak{}post-selection} represents the idealized baseline, whereas \textbf{QNLP:discard} denotes its practical counterpart. Both \textbf{QNLP}  models converge with a high accuracy across all three training, validation and test subsets with \textbf{post-selection} performing slightly better in training and validation, whereas \textbf{discard} does better in test.   \textbf{Q-BOW} is a baseline that ignores any semantic relationships across the text.
In the Q-BOW baseline, each word is first represented by a parameterised quantum state. The word states are then combined using a spider construction, the quantum implementation of the spider introduces CNOT operations to merge the individual word wires, while the trainable parameters reside in the word-state preparation circuits. 

As expected, it did not converge as well as the other QNLP models:  its training accuracy was around 70\%, its validation and test accuracies stagnated at around 58\% and 65\%. \textbf{BiLSTM@17M} reached similar accuracies to the QNLP models,  however it needed 17 million as opposed to the 47 thousand parameters. A parameter-matched \textbf{BiLSTM@47K}  did not train as well and its accuracies across all subsets were only in the mid 50 percents.

When training \textbf{Q-BOW},  the number of trainable parameters initially improved the generalization performance, with the mean test accuracy increasing from 58.38\% to approximately 77\% for models with around 17k–23k parameters. Beyond this point, however, further increasing the model size did not lead to additional improvements. Instead, both the mean test accuracy and training stability decreased, with the largest model exhibiting a lower mean test accuracy (67.11\%) and a higher standard deviation (10.90\%). 

\textbf{BiLSTM@47K} had an embedding dimension of 18 and hidden dimension of 36. With only one layer, the number of parameters reached 47K. When the layers were increased to two and three, the number of parameters increased to approximately 79K and 111K.  Increasing the embedding dimensions to 32 and the hidden dimensions to 64 increased the parameters to 1.6M even with one layer. Decreasing the dimensions to 16 and 32 decreased it to 43K. 

\section{Recommendation Generation and Results}

The recommender system utilizes an existing BBC Research and Development recommender package. We feed our learned pairwise similarity scores into this software for downstream recommendation generation. To obtain sufficiently broad similarity information across the programme catalogue, the held-out similarity predictions from all five cross-validation folds are combined, so that each labelled programme pair contributes an out-of-fold similarity estimate. These combined similarity scores are then used by the software to generate recommendations using an item-based $k$-nearest neighbours (KNN) model over a catalogue of 145 TV programmes. A single global recommender model is applied to each user’s viewing history to produce personalised recommendation lists. The two right-hand-side rectangles of Figure~\ref{quantum-pipeline-recommender} illustrate this procedure.

The evaluation set has 10,000 users.  Recommendations are produced for each of these users following the below steps: 

\begin{enumerate}
    \item Identify the set of programmes  watched by a user $u$.
    \item For each candidate programme, compute a relevance score by summing its similarities to the programmes in the user's history.
    \item Rank candidate programmes based on their scores.
    \item Recommend the top-$K$ highest-scoring programmes.
\end{enumerate}

We report results at cut off $K=20$, corresponding to the top-20 recommended items per user.  To evaluate the performance of the generated recommendation, we use both accuracy (MAP, NDCG) and diversity metrics (Surprisal, ILD, Coverage). The overall performance is obtained by averaging the results across all users. The definitions of these metrics are given below, following the
evaluation framework used in previous BBC recommender-system
work~\cite{CagaliSadrzadehNewell2021}.

\begin{enumerate}
\item \textbf{Mean Average Precision (MAP)} is defined as follows for $U$  the set of users in the evaluation dataset
\[
AP@K(u) = \frac{1}{\min(K, |G(u)|)} \sum_{n=1}^{K} P@n \cdot rel@n,
\]

\noindent
where $P@n$ is the precision at rank $n$, defined as the fraction of relevant items among the top-$n$ recommendations, and $rel@n = 1$ if the item at rank $n$ is in the ground truth set $G(u)$, and $0$ otherwise.
The overall MAP is computed as:
\[
MAP@K = \frac{1}{|U|} \sum_{u \in U} AP@K(u).
\]

\item \textbf{Normalised Discounted Cumulative Gain (NDCG)} is similar to MAP but additionally accounts for the rank positions of relevant programmes, giving greater importance to hits appearing higher in the recommendation list.
\accepted{
\[
\mathrm{DCG}@K(u)
=
\sum_{n=1}^{K}
\frac{\mathrm{rel}@n}{\log_2(n+1)},
~
\mathrm{NDCG}@K(u)
=
\frac{\mathrm{DCG}@K(u)}
{\mathrm{IDCG}@K(u)},
\]
where $\mathrm{IDCG}@K(u)$ denotes the DCG of the ideal ranking.}

\item \textbf{Intra-List Diversity (ILD)} \accepted{measures the diversity of genres within each user's recommendation list.
\[
\mathrm{ILD}@K(u)
=
\frac{2}{K(K-1)}
\sum_{i<j,\; i,j\in R_u}
\left(1-\cos(\mathbf{g}_i,\mathbf{g}_j)\right),
\]
where $R_u$ is the recommendation list for user $u$, and
$\mathbf{g}_i$ and $\mathbf{g}_j$ denote the genre representations
of programmes $i$ and $j$, respectively.}

\item \textbf{Surprisal} \accepted{reflects the presence of low-popularity programmes in the recommendations.
\[
I(i)=\log_2\!\left(\frac{|U|}{k_i}\right),
\qquad
\mathrm{Surprisal}@K(u)
=
\frac{1}{K}\sum_{i\in R_u} I(i),
\]
where $k_i$ is the number of users who consumed programme $i$ and
$|U|$ is the total number of users.}
\item \textbf{Coverage} \accepted{indicates the proportion of available programmes that appear in the recommendation lists.
\[
\mathrm{Coverage}@K
=
\frac{\left|\bigcup_{u\in U}R_u\right|}
{|I|},
\]
where $I$ denotes the full programme catalogue.}
\end{enumerate}
For user-level metrics, the reported score is obtained by averaging
the corresponding metric over all evaluated users.

Table~\ref{tab:not-hybrid} summarizes the performance of the evaluated recommender systems. Across the standalone configurations, ideal \textbf{QNLP:\allowbreak{}post-selection}, \textbf{QNLP:discard}, and the classical \textbf{BiLSTM} models demonstrated comparable results. 
In terms of MAP, QNLP:post-selection achieved the highest standalone score at 13.33\%, marginally outperforming BiLSTM@17M at 13.03\%, whereas the smaller BiLSTM@47K reached only 9.40\%.
Furthermore, \textbf{QNLP:\allowbreak{}post-selection} outperformed \textbf{QNLP:discard} (13.33\% vs. 12.93\%).

Following standard practice in recommendation ensembling~\cite{Kumar2025,grossi2022,waznisadrzadeh}, we constructed {hybrid} models by averaging the similarity scores of \textbf{QNLP:discard} with our classical content-based baselines \textbf{NTM-R}.
\accepted{Although the quantum model is trained using labels derived from NTM-R, it is only constrained to agree with NTM-R on the most similar and most dissimilar programme pairs selected for training. Intermediate pairs are not used for supervision, allowing the quantum model to generalise differently on these pairs and thereby provide complementary information when combined with NTM-R in the hybrid recommender.}
As shown in the  \emph{Hybrid} section of Table~\ref{tab:not-hybrid}, this combination outperformed all  of the stand alone  models and narrowed the performance gap with collaborative filtering (13.80\% vs. 16.00\% MAP). The combination \textbf{QNLP:post-selection + NTM-R} was our best hybrid model, reaching an MAP of 13.80\%, increasing the stand alone \textbf{post-selection} MAP by 0.47\%. The combination \textbf{QNLP:discard + NTM-R} had a similar increase (0.46\%), improving from an MAP of 12.93\% to 13.39\%. The \textbf{BiLSTM@17M + NTM-R} benefited from a similar increase (slightly lower and by 0.41\%). Combining all three models acted in a similar way, with \textbf{QNLP:post-selection+ BiLSTM + NTM-R} reaching a higher MAP than \textbf{QNLP:discard + BiLSTM + NTM-R}. We also experimented with combining the BiLSTM model with each of the QNLP ones; following the same trend the combination with post-selection did slightly better. \accepted{Finally, it is worth noting that due to the training complexity, the quantum model was only trained on short synopses of programmes. This is whereas its classical NTM-R version was trained on the full transcripts of the programmes. Despite this restriction, the ideal quantum model performed better than NTM-R and the practical one only under performed by 0.3\%. Generally, models such as NTM-R do not do well on short pieces of text and we believe that the quantum model excelled in the absence of data since it took the semantic structure of text into account.}

\section{Quantum Hardware Feasibility}\label{sec:quantum-runs}

The results in the preceding sections were obtained via classical
simulation: each circuit is evaluated by tensor network contraction,
and similarity is computed from the resulting representations. A
central motivation of this work is to demonstrate that these
computations can be carried out on quantum hardware. In this section,
we present a hardware feasibility study in which the trained circuits
are first executed on the Quantinuum H2 emulator~\cite{QuantinuumH2},
a classical emulation of the H2 trapped-ion quantum
processor~\cite{MosesH2} that reproduces its gate set, connectivity,
and noise characteristics, and then on the physical H2-2 device. We
address two questions: (i)~whether the similarity signal encoded in
the circuits is recoverable via a hardware-compatible measurement
protocol, and (ii)~whether emulator results reliably predict hardware
performance on our task.

We evaluate 99 programme pairs from the validation split. This subset retains the label balance of the full dataset and is small enough to run exhaustively on the emulator. We emphasise that this evaluation is distinct from the 5-fold cross-validation reported in Table~\ref{tab:pairwise_similarity}: the cross-validation assesses model generalisation across the full dataset, while the hardware experiments assess whether the learned similarity is recoverable on a quantum device for a representative subset.

\subsection{The Quantinuum H2 System and Emulator}\label{sec:h2-system}

The Quantinuum H2 is a trapped-ion quantum processor based on a race-track architecture~\cite{MosesH2}, with the following properties:

\begin{itemize}
    \item \textbf{All-to-all connectivity}: any pair of qubits can interact directly, without SWAP routing. This is important for our protocol, which applies gates between qubits in different circuit registers.
    \item \textbf{High-fidelity gates}: two-qubit gate fidelities exceed $99.5\%$~\cite{QuantinuumH2}, with a native gate set of single-qubit rotations, two-qubit ZZ gates, parameterized-angle ZZ gates,
general SU(4) entangler.
    \item \textbf{Up to 56 qubits}: our combined circuits \accepted{for each synopsis pair} range from 9 to 23 qubits, well within capacity.
\end{itemize}

The H2 emulator~\cite{QuantinuumH2} reproduces the device's gate set and compilation passes, and supports two modes:
\begin{itemize}
    \item \textbf{Noiseless}: ideal unitary evolution with sampling noise only (standard error $1/\sqrt{N}$ for $N$ shots).
    \item \textbf{Noisy}: includes a noise model reproducing the characteristics of the physical device.
\end{itemize}

Running both modes allows us to separate two effects: whether the measurement protocol is correct (noiseless mode) and how device noise degrades the signal (noisy mode).

\subsection{Why a Quantum Measurement Protocol is Needed}\label{sec:why-swap}

When a single circuit is executed on hardware, the result is a set of bitstrings sampled from a probability distribution. These bitstrings give access to the Born-rule probabilities $|\alpha_i|^2$, but not to the complex amplitudes $\alpha_i$ themselves. 
The similarity metric used in training is the Uhlmann fidelity $F(\rho_A,\rho_B)$ between the reduced output states. This quantity cannot in general be reconstructed from the individual measurement statistics of circuits $A$ and $B$ separately.

To compute the state overlap without reconstructing the states, we require a protocol that compares the two circuits \emph{jointly} at the quantum level. The destructive SWAP test provides this capability.

In our circuits, the situation is further complicated by the presence of discarded qubits. On hardware, all qubits are physically measured, but the discarded wires' outcomes are not used. Because the entangling gates create correlations between the discarded and non-discarded qubits, the marginal statistics on the output qubits alone correspond to those of a mixed state.

\subsection{The Destructive SWAP Test}\label{sec:swap-test}

We employ the destructive SWAP test~\cite{BuhrmanCleveWigderson2001, Cincio2018} to estimate the state overlap 
\begin{equation}\label{eq:state-overlap}
    \mathrm{Tr}(\rho_A \rho_B).
\end{equation}
This quantity compares the marginal statistics of the output qubits between two circuits. For pure states, it reduces to $|\langle\psi_A|\psi_B\rangle|^2$. For our case, it does not reduce to the latter formula since we are not dealing with pure states.

\paragraph{Protocol.}
Given circuits $A$ ($n_A$ qubits) and $B$ ($n_B$ qubits), each producing $n_{\mathrm{out}} = 2$ output qubits:
\begin{enumerate}
    \item Run circuit $A$ on qubits $\{0, \ldots, n_A{-}1\}$ and circuit $B$ on qubits $\{n_A, \ldots, n_A{+}n_B{-}1\}$. The two circuits use \emph{disjoint} qubit registers initialised from $\ket{0}$ .
    \item Apply a Bell-basis measurement across each pair of corresponding output qubits:
    \begin{equation}
        \mathrm{CNOT}(q^A_i \to q^B_i), \quad H(q^A_i), \quad \text{measure both},
    \end{equation}
    for $i \in \{0, 1\}$.
    \item Measure all remaining qubits in the computational basis (required by the H2 system). 
    \item The hardware returns a set of bitstrings and their counts. From each bitstring, extract the four bits at the Bell-measured positions ($z_1, z_2$ from the Hadamard-transformed output qubits of $A$, and $x_1, x_2$ from the CNOT targets in $B$). Compute:
    \begin{equation}\label{eq:swap-estimator}
        \mathrm{Tr}(\rho_A \rho_B) = \frac{1}{N}\sum_{k=1}^{N}\prod_{i=1}^{n_{\mathrm{out}}} (-1)^{z_i^{(k)} \cdot x_i^{(k)}},
    \end{equation}
    where $k$ indexes the shots. Each shot contributes $+1$ or $-1$; the average over $N$ shots converges to $\mathrm{Tr}(\rho_A \rho_B)$.
\end{enumerate}

The overhead is minimal: $n_{\mathrm{out}}$ additional CNOT gates and $n_{\mathrm{out}}$ Hadamard gates, with no ancilla qubits. The total qubit count is $n_A + n_B$. On the H2 system, all-to-all connectivity ensures the cross-register gates incur no routing overhead.

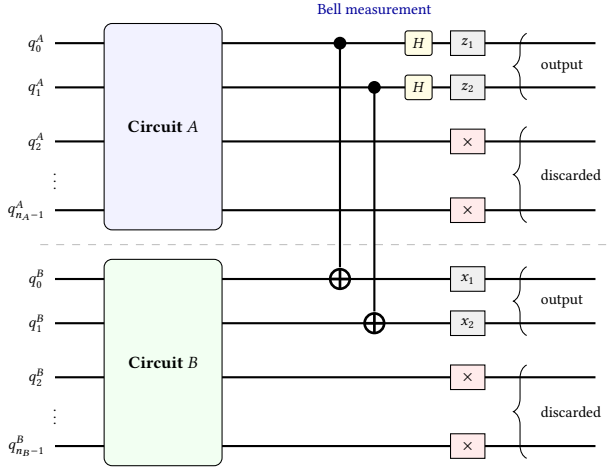
\begin{figure}[t]
\centering
\begin{tikzpicture}[
    scale=0.65, transform shape,
    gate/.style={draw, fill=blue!8, minimum width=0.9cm, minimum height=0.5cm,
                 font=\normalsize, rounded corners=1pt},
    meas/.style={draw, fill=gray!12, minimum width=0.7cm, minimum height=0.5cm,
                 font=\normalsize},
    disc/.style={draw, fill=red!8, minimum width=0.7cm, minimum height=0.5cm,
                 font=\normalsize},
    wire/.style={thick},
    brace/.style={decorate, decoration={brace, amplitude=5pt, mirror}},
    cnottarget/.style={draw, circle, inner sep=4pt, line width=1pt},
]
\draw[wire] (0, 0) -- (11, 0);
\draw[wire] (0, -0.9) -- (11, -0.9);
\draw[wire] (0, -2.0) -- (11, -2.0);
\node[font=\normalsize] at (0, -2.7) {$\vdots$};
\draw[wire] (0, -3.4) -- (11, -3.4);
\draw[wire] (0, -4.8) -- (11, -4.8);
\draw[wire] (0, -5.7) -- (11, -5.7);
\draw[wire] (0, -6.8) -- (11, -6.8);
\node[font=\normalsize] at (0, -7.5) {$\vdots$};
\draw[wire] (0, -8.2) -- (11, -8.2);
\node[anchor=east, font=\normalsize] at (-0.1, 0) {$q^A_0$};
\node[anchor=east, font=\normalsize] at (-0.1, -0.9) {$q^A_1$};
\node[anchor=east, font=\normalsize] at (-0.1, -2.0) {$q^A_2$};
\node[anchor=east, font=\normalsize] at (-0.1, -3.4) {$q^A_{n_A{-}1}$};
\node[anchor=east, font=\normalsize] at (-0.1, -4.8) {$q^B_0$};
\node[anchor=east, font=\normalsize] at (-0.1, -5.7) {$q^B_1$};
\node[anchor=east, font=\normalsize] at (-0.1, -6.8) {$q^B_2$};
\node[anchor=east, font=\normalsize] at (-0.1, -8.2) {$q^B_{n_B{-}1}$};
\draw[dashed, gray!50] (-0.3, -4.1) -- (11.3, -4.1);
\node[draw, fill=blue!5, minimum width=2.4cm, minimum height=4.2cm,
      rounded corners=3pt, font=\large\bfseries] at (2.2, -1.7) {Circuit $A$};
\node[draw, fill=green!5, minimum width=2.4cm, minimum height=4.2cm,
      rounded corners=3pt, font=\large\bfseries] at (2.2, -6.5) {Circuit $B$};
\fill (5.8, 0) circle (0.13);
\draw[thick] (5.8, 0) -- (5.8, -4.56);
\node[cnottarget] (cnot1) at (5.8, -4.8) {};
\draw[line width=1pt] (cnot1.west) -- (cnot1.east);
\draw[line width=1pt] (cnot1.north) -- (cnot1.south);
\fill (6.5, -0.9) circle (0.13);
\draw[thick] (6.5, -0.9) -- (6.5, -5.46);
\node[cnottarget] (cnot2) at (6.5, -5.7) {};
\draw[line width=1pt] (cnot2.west) -- (cnot2.east);
\draw[line width=1pt] (cnot2.north) -- (cnot2.south);
\node[gate, fill=yellow!12, minimum width=0.55cm] at (7.4, 0) {$H$};
\node[gate, fill=yellow!12, minimum width=0.55cm] at (7.4, -0.9) {$H$};
\node[meas] at (8.4, 0) {$z_1$};
\node[meas] at (8.4, -0.9) {$z_2$};
\node[meas] at (8.4, -4.8) {$x_1$};
\node[meas] at (8.4, -5.7) {$x_2$};
\node[disc] at (8.4, -2.0) {$\times$};
\node[disc] at (8.4, -3.4) {$\times$};
\node[disc] at (8.4, -6.8) {$\times$};
\node[disc] at (8.4, -8.2) {$\times$};
\draw[brace] (9.6, 0.2) -- (9.6, -1.15)
    node[midway, right=5pt, font=\normalsize] {output};
\draw[brace] (9.6, -1.7) -- (9.6, -3.65)
    node[midway, right=5pt, font=\normalsize] {discarded};
\draw[brace] (9.6, -4.55) -- (9.6, -5.95)
    node[midway, right=5pt, font=\normalsize] {output};
\draw[brace] (9.6, -6.55) -- (9.6, -8.45)
    node[midway, right=5pt, font=\normalsize] {discarded};
\node[font=\normalsize, text=blue!60!black] at (6.5, 0.7) {Bell measurement};
\end{tikzpicture}
\caption{Destructive SWAP test. Circuits $A$ and $B$ run forward on disjoint registers. A Bell-basis measurement (CNOT + $H$) is applied across the output qubit pairs. Discarded qubits ($\times$) are measured but not used.}
\label{fig:swap-circuit}
\end{figure}

\subsection{Does the Similarity Signal Survive?}\label{sec:signal-survival}


The circuits were trained to maximise the Uhlmann fidelity: \[
F(\rho_a, \rho_b) = \left( \text{Tr} \sqrt{\rho_a^{1/2} \rho_b \rho_a^{1/2}} \right)^2.
\] for similar programmes and minimise it for dissimilar ones. On hardware, we measure $\mathrm{Tr}(\rho_A \rho_B)$  via the SWAP test. 

We tested if the similarity signal survives in our protocol by computing $\mathrm{Tr}(\rho_A \rho_B)$ exactly for all 99 pairs via full statevector simulation and numerical partial trace, and comparing against the classical training fidelity. Table~\ref{tab:dm-validation} shows the average values for each class. The Pearson correlation between the two metrics is $r = 0.910$. 

\begin{table}[!h]
\centering
\renewcommand{\arraystretch}{1.3}
\setlength{\tabcolsep}{6pt}
\begin{tabular}{l c c}
\toprule
\textbf{Metric} & \textbf{Dissimilar} & \textbf{Similar} \\
\midrule
Training fidelity (Uhlmann)
& $0.44 \pm 0.19$
& $0.85 \pm 0.10$ \\
SWAP test overlap 
& $0.174 \pm 0.042$
& $0.368 \pm 0.065$ \\
\bottomrule
\end{tabular}
\caption{ Comparison of the classical training fidelity and the SWAP test overlap (noiseless emulator, 1000 shots) on 99 programme pairs (46 dissimilar, 53 similar). There is a strong correlation $r = 0.910$ between the two, confirming that the similarity signal is preserved in the output-qubit state overlap.}
\label{tab:dm-validation}
\end{table}

The training fidelity (Uhlmann) spans $[0, 1]$, while the
hardware-measurable overlap $\mathrm{Tr}(\rho_A \rho_B)$ occupies a
compressed range. For two $d$-dimensional quantum states,
$\mathrm{Tr}(\rho \sigma) \leq \mathrm{Tr}(\rho^2)$ with equality
when $\rho = \sigma$, and
$\mathrm{Tr}(\rho \sigma) \geq 1/d^2$ for states drawn uniformly at
random~\cite{NielsenChuang}. For our two-qubit outputs ($d = 4$), the
lower bound is $1/16 = 0.0625$. Since discarding entangled qubits
generically produces mixed states, the purities of our circuit outputs
are substantially less than~1 (mean purity $\approx 0.36$ across the
dataset). This compresses the effective range of
$\mathrm{Tr}(\rho_A \rho_B)$ to approximately $[0.06, 0.50]$ on the
emulator and $[0.08, 0.53]$ on hardware, compared to $[0, 1]$ for the
training fidelity. The same compression is observed on both emulator
and hardware because it is a property of the quantum states, not of
the measurement device.
 
Crucially, this compression is monotonic: the ranking of pairs by
$\mathrm{Tr}(\rho_A \rho_B)$ preserves the ranking by training
fidelity ($r = 0.910$ on the full emulator dataset, $r = 0.963$ on
the hardware subset). To obtain a classification threshold in the overlap space, we
project the training threshold of $0.5$ (in Uhlmann fidelity)
through the linear relationship between fidelity and overlap,
yielding a mapped threshold of $0.213$. This mapping uses only
the continuous-valued fidelity and overlap scores, not the
binary labels; the threshold of $0.5$ was fixed during
classical training before any quantum measurement was
performed. \accepted{As a consistency check, a threshold derived purely
from the theoretical bounds (the midpoint of the accessible
range $\frac{1}{2}(1/d^2 + \bar{p}) \approx 0.21$, where
$\bar{p} \approx 0.36$ is the mean purity and $d = 4$ is the
output dimension) yields a comparable value without
reference to any data.}
At this threshold, the classification accuracy is $91.9\%$
(noiseless, 1000 shots) and $88.9\%$ (noisy, 1000 shots).

\subsection{Emulator Results}\label{sec:emulator-results}

We executed the destructive SWAP test on the H2 emulator in both noiseless and noisy modes at 100 and 1000 shots per pair. After compilation to the H2 native gate set with optimisation level 2, the combined circuits ranged from 9 to 23 qubits, 124 to 676 compiled
two-qubit gates, and 44 to 150 two-qubit gate depth.

\subsubsection{Sampling Noise}

\begin{figure}[!h]
    \centering
    \includegraphics[width=\linewidth]{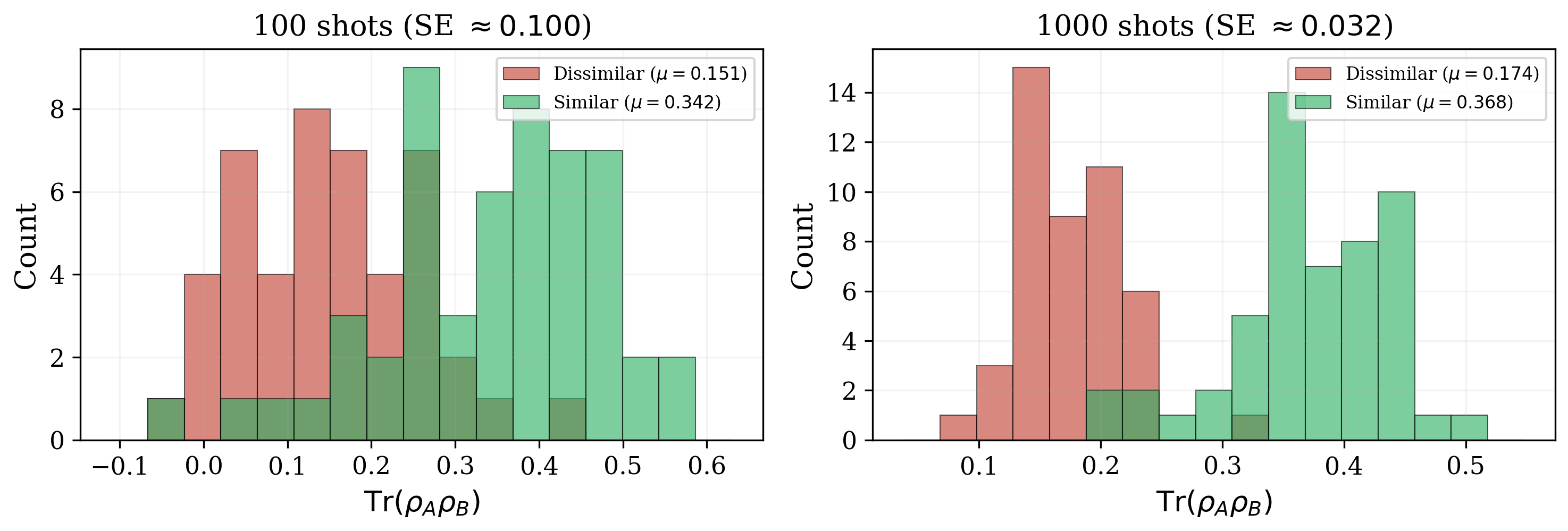}
\caption{Effect of shot budget on the distribution of SWAP test overlaps. Left: 100 shots ($\mathrm{SE} \approx 0.10$), where the distributions overlap. Right: 1000 shots ($\mathrm{SE} \approx 0.03$), where the distributions are more separated. Since $\mathrm{SE} = 1/\sqrt{N}$, increasing the shot count improves the approximation of the true overlap at the cost of longer runtimes.}
    \label{fig:shot-budget}
\end{figure}

Each shot of the SWAP test produces a value of $+1$ or $-1$ via the parity formula~\eqref{eq:swap-estimator}. The overlap estimate is the average over $N$ such values. Since each value is bounded by $\pm 1$, the variance per shot is at most 1, and the standard error of the mean is
\begin{equation}
    \mathrm{SE} = \frac{1}{\sqrt{N}}.
\end{equation}
At $N = 1000$ shots, $\mathrm{SE} \approx 0.032$. The average SWAP test overlap for dissimilar pairs is $0.174$ and for similar pairs $0.368$, a gap of approximately $0.194$. With $\mathrm{SE} \approx 0.032$, this gap is roughly $6.1$ standard errors wide. For any individual pair, the measurement is noisy, but across the full population of pairs, the group averages are reliably separated. At 100 shots ($\mathrm{SE} \approx 0.10$), the gap of $0.194$ remains approximately $1.9$ standard errors wide, providing meaningful separation even at lower shot counts (Figure~\ref{fig:shot-budget}).

\subsubsection{Correlation with Classical Training Fidelity}

\begin{figure}[!h]
    \centering
    \includegraphics[width=0.75\linewidth]{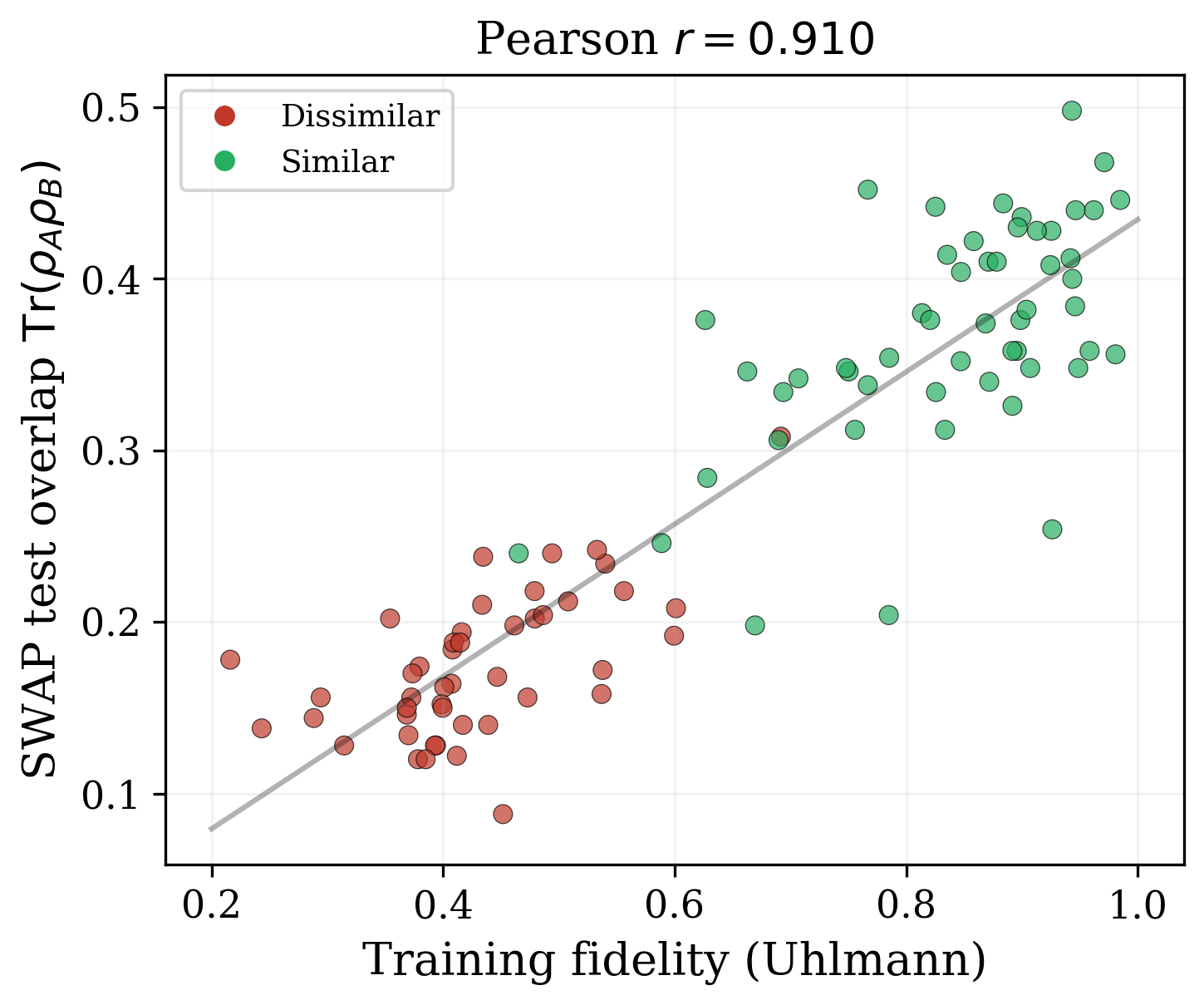}
    \caption{Training fidelity (Uhlmann) versus emulator SWAP test overlap (noiseless, 1000 shots). The correlation $r = 0.910$ confirms that the similarity signal is recoverable from sampled measurements. Red: dissimilar; green: similar.}
    \label{fig:swap-vs-classical}
\end{figure}

Figure~\ref{fig:swap-vs-classical} shows the emulator SWAP test overlap (noiseless, 1000 shots) versus the training fidelity (Uhlmann). The Pearson correlation is $r = 0.910$. Despite this, the overall trend is clear and the red/green separation is visible, indicating that the similarity signal learned during training is recoverable from hardware measurements.

\begin{figure}[!h]
    \centering
    \includegraphics[width=\linewidth]{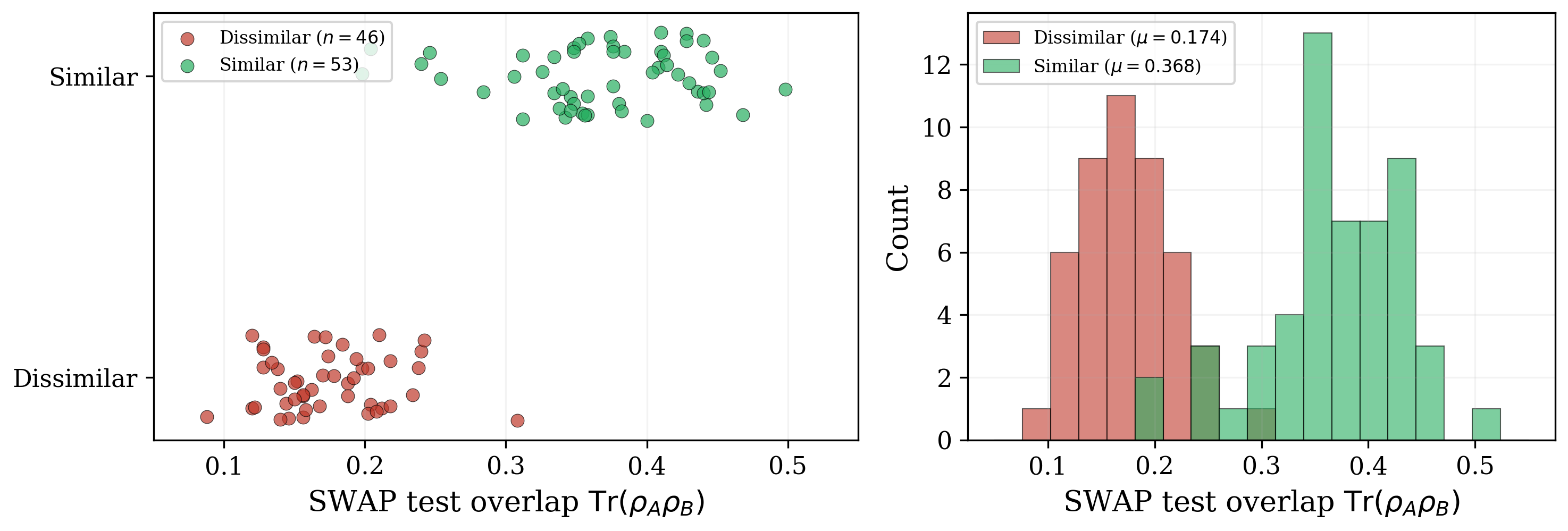}
    \caption{Distribution of SWAP test overlaps on the H2 emulator (noiseless, 1000 shots) by label. Left: strip plot showing individual pairs. Right: histogram with group means. Dissimilar pairs cluster around $0.174$ and similar pairs around $0.368$.}
    \label{fig:overlap-by-label}
\end{figure}

Figure~\ref{fig:overlap-by-label} shows the distribution of SWAP test overlaps by label on the emulator (noiseless, 1000 shots). The dissimilar pairs cluster around $0.174$ and the similar pairs around $0.368$, with partial overlap between the distributions due to sampling noise. As noted above, the standard error scales as $1/\sqrt{N}$, so higher shot counts would further sharpen the separation at the cost of increased runtime.

\subsubsection{Effect of Device Noise}

\begin{figure}[!h]
    \centering
    \includegraphics[width=0.75\linewidth]{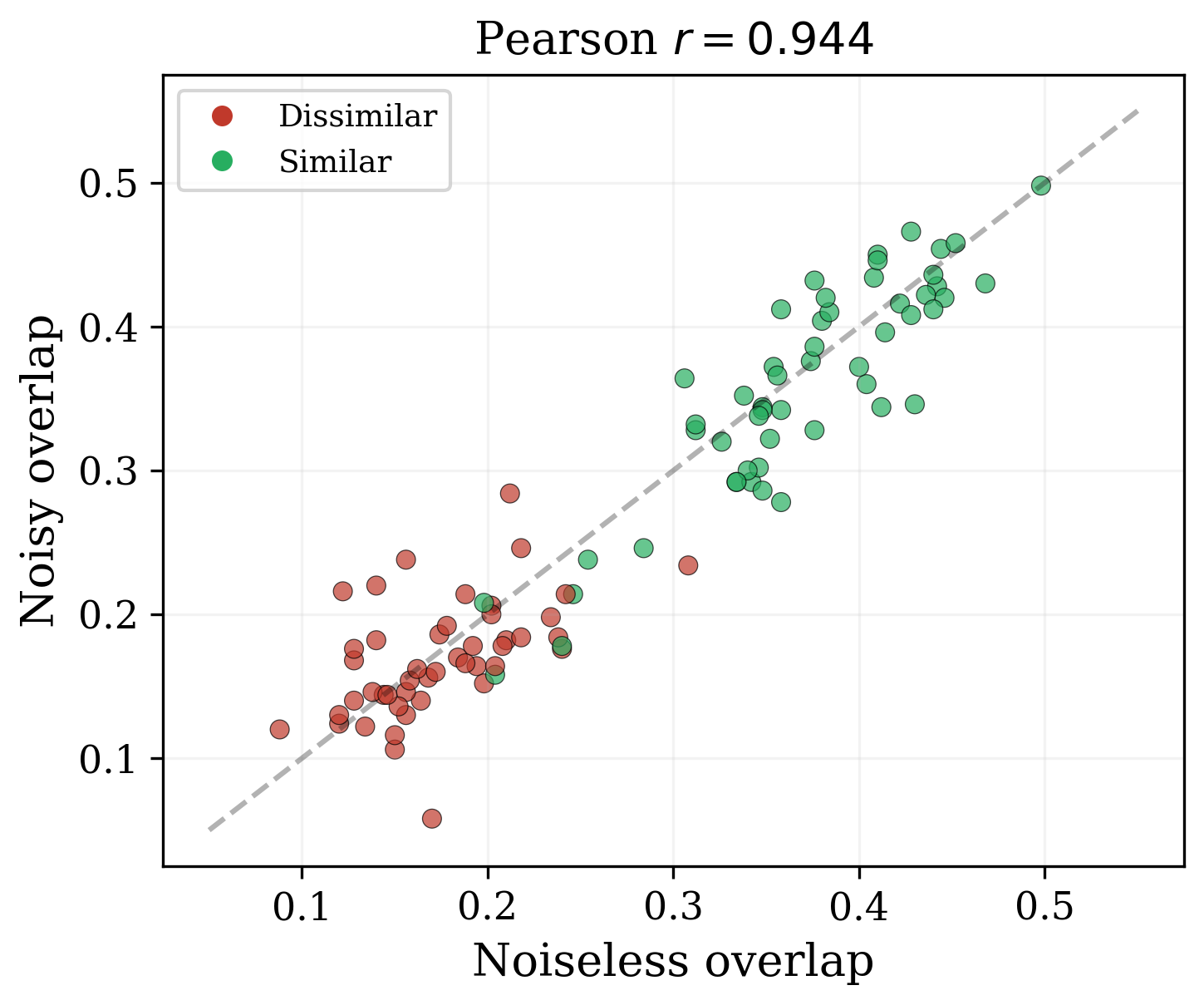}
    \caption{Noisy versus noiseless emulator overlap at 1000 shots ($r = 0.944$). Points scatter around the diagonal, indicating that the noise model adds per-pair variance but does not systematically degrade the signal.}
    \label{fig:noisy-vs-noiseless}
\end{figure}

Figure~\ref{fig:noisy-vs-noiseless} compares noiseless and noisy emulator results across all 99 pairs. In aggregate, the effect of the noise model is small: the group means are nearly identical dissimilar: $0.174$ noiseless vs $0.170$ noisy; similar: $0.368$ vs $0.358$), and the correlation between the two is $r = 0.944$). The noise model introduces additional per-pair variance beyond sampling noise, but does not systematically bias the overlap in one direction. Table~\ref{tab:emulator-summary} summarises the aggregate results.

\begin{figure}[t]
    \centering
    \includegraphics[width=0.75\linewidth]{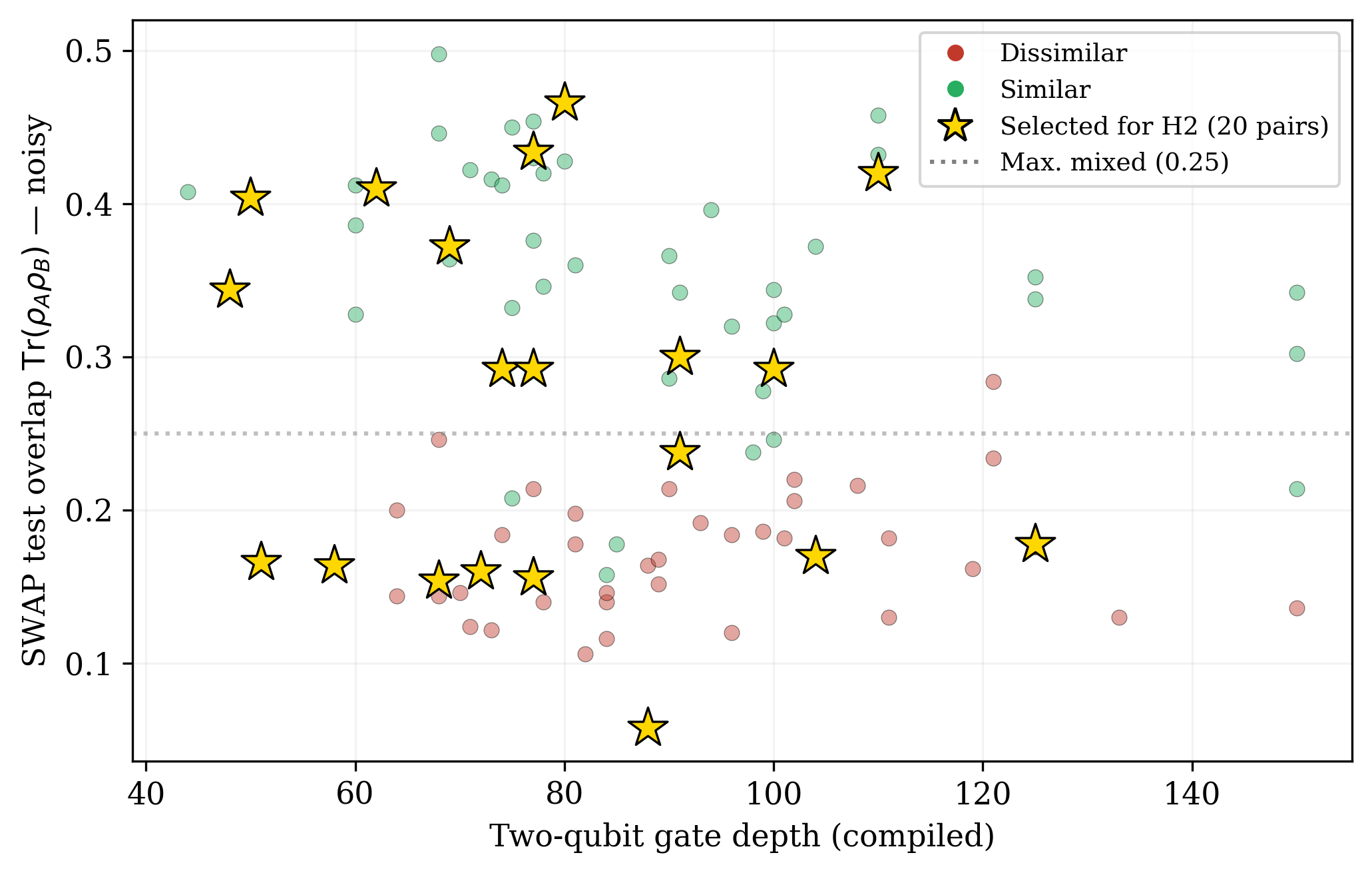}
    \caption{Emulator SWAP test overlap (noisy, 1000 shots) versus compiled two-qubit gate depth. For circuits below approximately depth 100, the separation between similar and dissimilar pairs is clear. For the deepest circuits (depth $>120$), both classes are compressed toward $0.25$, the overlap value expected for maximally mixed states.}
    \label{fig:overlap-vs-gates}
\end{figure}

We also show the noisy SWAP test overlap as a function of the compiled two-qubit gate depth (Figure~\ref{fig:overlap-vs-gates}).

\begin{table}[t]
\centering
\caption{Emulator SWAP test results at 1000 and 100 shots. Accuracy
is computed at the mapped training threshold ($0.213$ for noiseless,
$0.207$ for noisy, $0.184$ for 100 shots), obtained by projecting the
Uhlmann fidelity threshold of $0.5$ into the overlap space via linear
regression.}
\label{tab:emulator-summary}
\begin{tabular}{l c c c c}
\toprule
\textbf{Configuration} & \textbf{Dissimilar} & \textbf{Similar} & $r$ & \textbf{Acc.} \\
\midrule
Noiseless, 1000 shots  & $0.174 \pm 0.042$ & $0.368 \pm 0.065$ & $0.910$ & $91.9\%$ \\
Noisy, 1000 shots      & $0.170 \pm 0.041$ & $0.358 \pm 0.077$ & $0.875$ & $88.9\%$ \\
Noiseless, 100 shots   & $0.151 \pm 0.101$ & $0.342 \pm 0.129$ & $0.690$ & $79.8\%$ \\
\bottomrule
\end{tabular}
\end{table}

\subsection{H2 Hardware Results}

Running all 99 emulator pairs on the H2 processor at 1000 shots would
require more than the
allocated hardware budget for this project. Therefore, we selected a representative subset of
20~circuit pairs that spans the full range of circuit sizes in
the dataset. The selection covers four qubit-count bins (12--14, 15--17,
18--20, 21--23) and four compiled two-qubit gate depth bins (44--70,
71--90, 91--110, 111--150), with both similar and dissimilar pairs in
each populated bin. The subset preserves the statistical properties of
the full dataset: the mean qubit count (15.7 vs 16.3), mean two-qubit
gate depth (84 vs 89), and the emulator overlap gap on the selected subset ($0.238$) is
comparable to the full-dataset gap ($0.194$), with the difference
attributable to the subset's bias toward smaller, better-separated
circuits.
We executed the circuits on the Quantinuum H2-2 trapped-ion
processor for this representative subset of 20 circuit pairs at 1000 shots per pair.

Table~\ref{tab:hw-results} summarises the hardware results alongside the
noisy emulator results for the same 20~pairs.
 
\begin{table}[t]
\centering
\caption{Hardware (H2-2) and noisy emulator results for the 20 selected
pairs at 1000 shots. The hardware gap ($0.257$) exceeds the emulator
prediction ($0.238$).}
\label{tab:hw-results}
\begin{tabular}{l c c}
\toprule
& \textbf{H2-2 Hardware} & \textbf{Noisy Emulator} \\
\midrule
Dissimilar ($n = 9$)  & $0.157 \pm 0.042$ & $0.158 \pm 0.044$ \\
Similar ($n = 11$)    & $0.414 \pm 0.058$ & $0.395 \pm 0.061$ \\
Gap                   & $0.257$           & $0.238$ \\
Cohen's $d$           & $5.11$            & $4.48$ \\
$r$ with training fidelity & $0.963$      & --- \\
\bottomrule
\end{tabular}
\end{table}
 
\begin{figure}[t]
\centering
\includegraphics[width=0.75\linewidth]{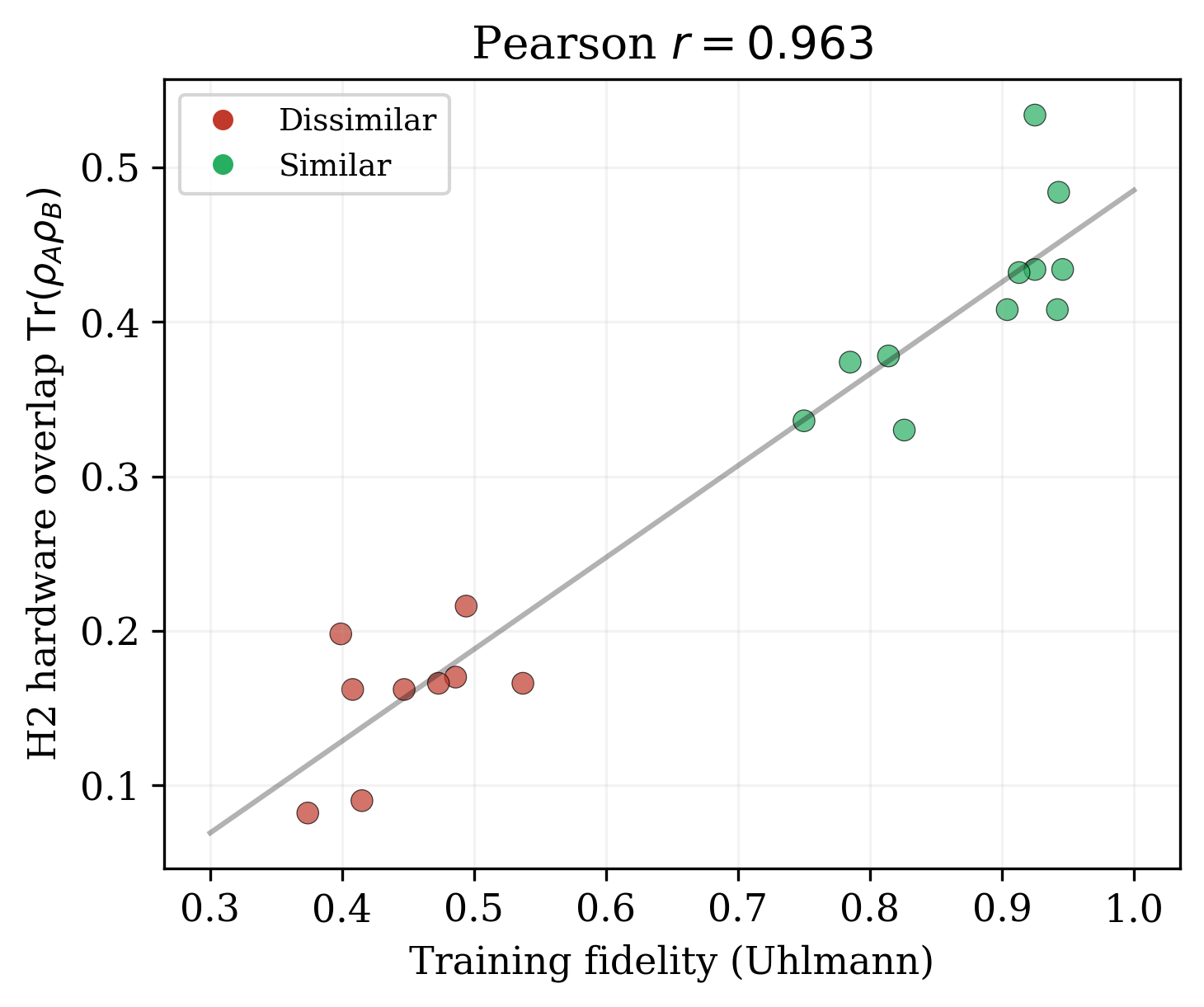}
\caption{H2-2 hardware overlap versus training fidelity (Uhlmann),
$r = 0.963$. Circuits trained to have high similarity produce high
state overlap on hardware.}
\label{fig:hw-vs-fidelity}
\end{figure}

The similarity signal survives execution on quantum hardware. Dissimilar
pairs produce a mean overlap of $0.157 \pm 0.042$ and similar pairs
$0.414 \pm 0.058$, yielding a gap of $0.257$ (Cohen's $d = 5.11$,
$p = 3.5 \times 10^{-9}$). The hardware overlap correlates with the
training fidelity at $r = 0.963$ (Figure~\ref{fig:hw-vs-fidelity}),
confirming that circuits trained to have high similarity also produce
high state overlap on hardware.
 
\begin{figure}[t]
\centering
\includegraphics[width=0.75\linewidth]{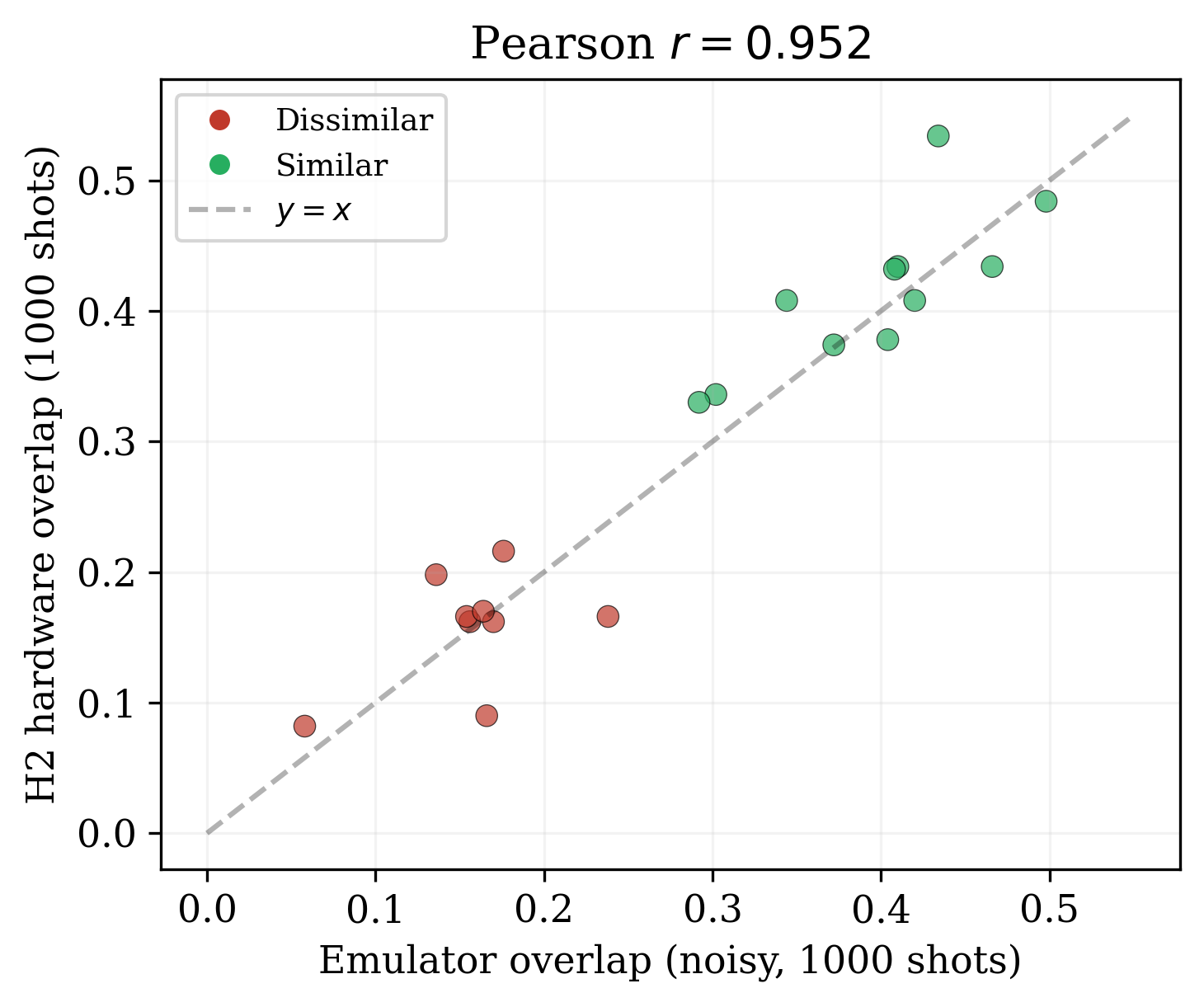}
\caption{H2-2 hardware overlap versus noisy emulator overlap for the
same 20 pairs ($r = 0.952$). Points near the diagonal confirm the
emulator is a reliable predictor of hardware performance.}
\label{fig:hw-vs-emulator}
\end{figure}

Figure~\ref{fig:hw-vs-emulator} compares the hardware and noisy emulator
overlaps for each pair. The Pearson correlation is $r = 0.952$, and the
points cluster around the diagonal, confirming that the noisy emulator
is a reliable predictor of hardware performance. The hardware gap
($0.257$) is slightly larger than the emulator prediction ($0.238$),
indicating that the emulator noise model provides a conservative
estimate.
 
\begin{figure}[t]
\centering
\includegraphics[width=0.85\linewidth]{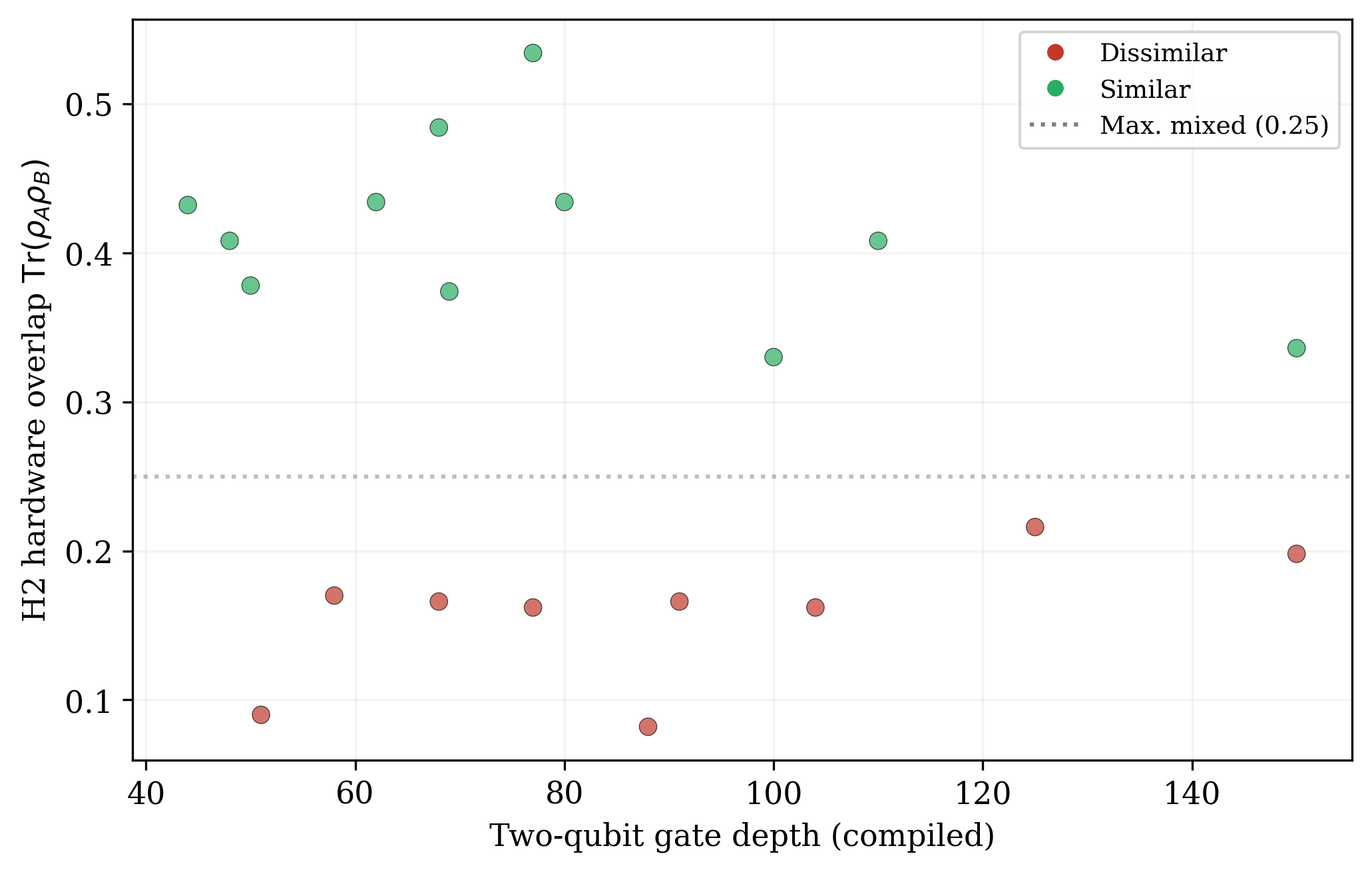}
\caption{H2-2 hardware overlap versus compiled two-qubit gate depth.
The separation between similar and dissimilar pairs is maintained across
all depths, including the deepest circuits (depth 150).}
\label{fig:hw-vs-depth}
\end{figure}

\begin{figure}[t]
\centering
\includegraphics[width=0.85\linewidth]{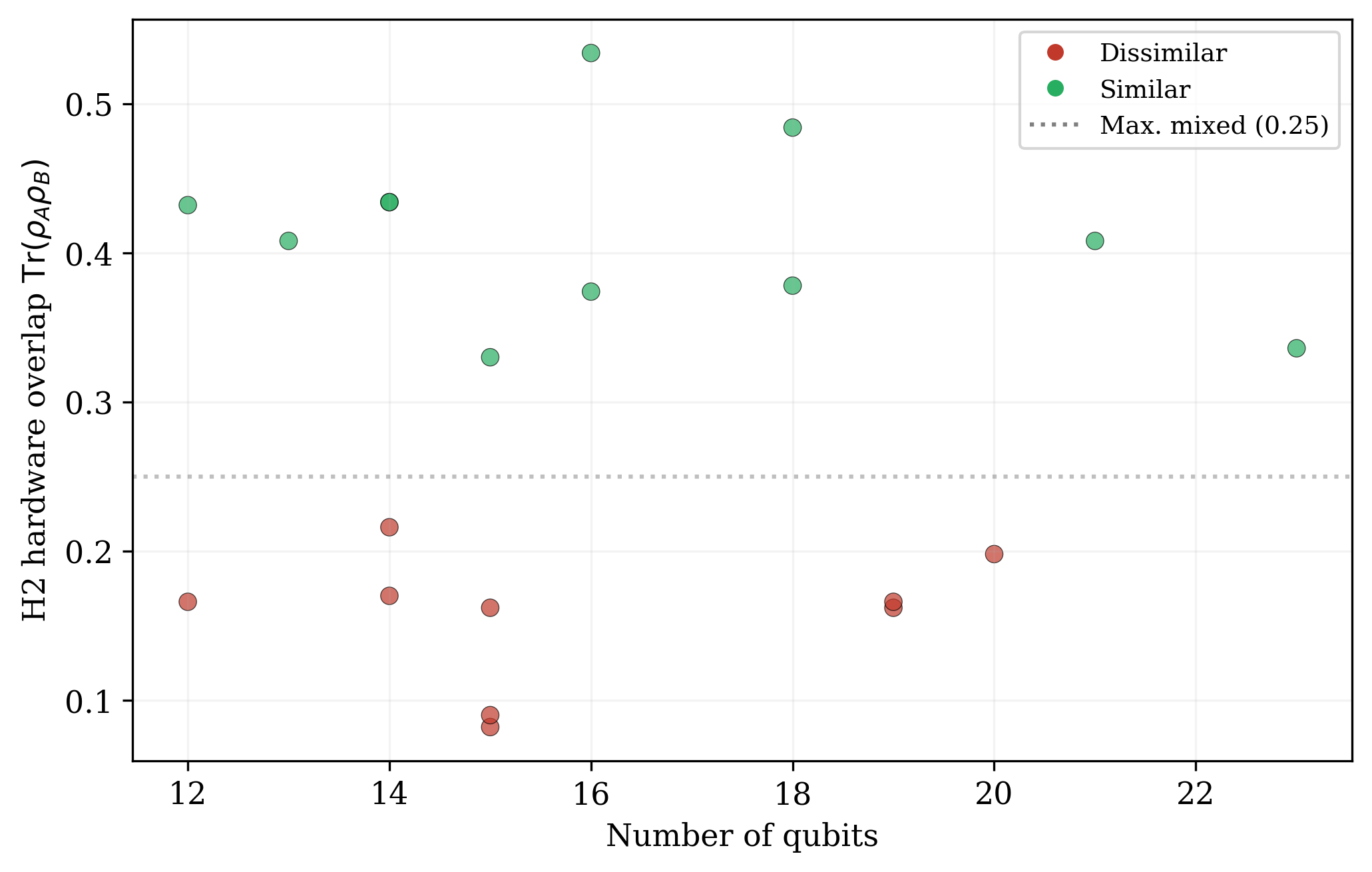}
\caption{H2-2 hardware overlap versus number of qubits. The 
separation between similar and dissimilar pairs is maintained 
across all circuit sizes (12 to 23 qubits). The dashed line 
shows the maximally mixed baseline (0.25).}
\label{fig:hw-vs-qubits}
\end{figure}

Figure~\ref{fig:hw-vs-depth} shows the hardware overlap as a function
of the compiled two-qubit gate depth. The separation between similar
and dissimilar pairs is maintained across all depths tested, including
circuits at depth~150. As expected, the deepest circuits show the most
variance, but the signal remains above the maximally mixed baseline. Figure~\ref{fig:hw-vs-qubits} shows the same trend as a function of qubit count, with separation maintained from 12 to 23 qubits.

\begin{figure}[t]
\centering
\includegraphics[width=\linewidth]{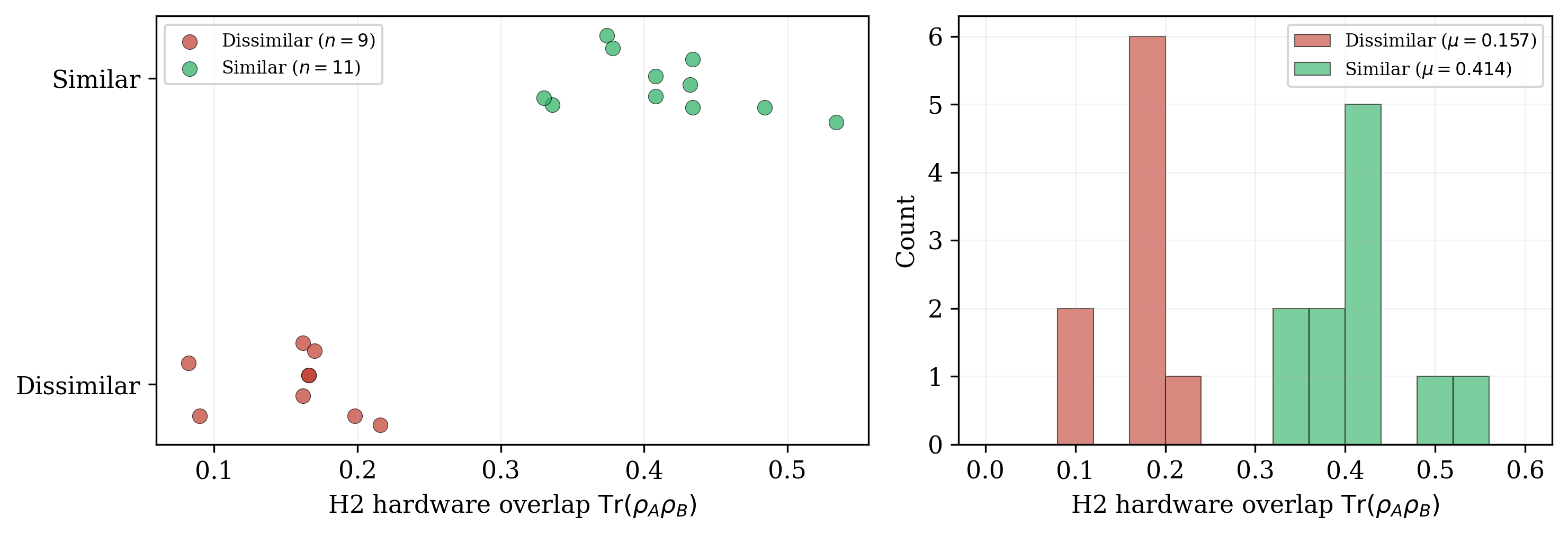}
\caption{H2-2 hardware results (1000 shots, 20 pairs). Left: strip
plot showing clear separation between dissimilar (red) and similar
(green) pairs. Right: histogram of hardware overlaps
($\mu_{\text{dis}} = 0.157$, $\mu_{\text{sim}} = 0.414$, gap $= 0.257$).}
\label{fig:hw-main}
\end{figure}
 
\section{Summary and Next Steps}\label{sec:hw-summary}
The work presented in this paper demonstrated that a quantum content-based recommender system performs on par with a classical BiLSTM one, with a MAP difference of 0.10 (the BiLSTM MAP was 13.03, and the quantum one 12.93), but with significantly less parameters: 47K vs 17.8M. A parameter-matched BiLSTM decreased the  MAP by 3.63. It went down from 13.03 to 9.40.  When combined with  the classical NTM-R content-based recommender models MAP increased by 0.46\%-0.47\% across the two QNLP models. This confirms previous results in the same field, where (1) combining different content-based models proved to be useful \cite{CagaliSadrzadehNewell2021,NazirCagaliSadrzadehNewell2020}, and (2) combining classical and quantum results resulted in a boost in accuracy \cite{Kumar2025,grossi2022,waznisadrzadeh}. Our main quantum models followed a framework known as \textbf{QNLP}, which embeds the semantic structure of the text into variational quantum circuits. The performance of this model was far better than an unstructured model that embedded a bag-of-words model of meaning into circuits (an MAP of 9.30\% vs the ideal 13.33\% and the practical 12.93\%). This shows  firstly, semantic structure plays a major role in a content-based recommender system, and secondly, quantum circuits are able to model this structure. 

In this paper we focused on content, extracted from the textual assets of programmes. In future work, we will experiment with other   signals such as genre and audiovisual assets. With regards to genre,  each programme has a  tree describing to which hierarchy of categories does it belong, e.g. documentary-nature\&nature, or documentary-history. Our roadmap would be to first train our QNLP models with the genre similarity labels, then combine these with the original genre model, and finally combine these ensembles with the \accepted{NTM-R} models, allowing for a three-way QNLP+\accepted{NTM-R}+genre hybrid recommender. Audiovisual assets come from the programmes audio and video files, which can be modelled into embedding vectors in ways similar to text. In previous work, they led to small increases in accuracy but  required far more compute~~\cite{CagaliSadrzadehNewell2021,NazirCagaliSadrzadehNewell2020}. We are hoping quantum-inspired approaches such as the one used in this paper, will improve on this. 

A potential shortcoming of the current model is its use of a single merge box that combines all of the words within a piece of text. It is not optimal to use the same merge box  for new pieces of text, especially when the size of their text-circuit is wider than 
the ones already seen in the current dataset. A solution to this problem is the use of a series of binary merge boxes in a staircase pattern. In this case, the binary box merges two words at the time and can be applied to a text of any length. Implementing this model is left to future work.

The model was run on the hardware emulator on Quantinuum's H2 device. The results were used to provide a  hardware feasibility study for this task. This established three results:

\begin{enumerate}
  \item \textbf{The signal survives on quantum hardware.} On 20
    representative circuit pairs executed on the H2-2 processor, the
    SWAP test overlap separates similar from dissimilar pairs with a gap
    of $0.257$ (Cohen's $d = 5.11$). The hardware overlap correlates
    with the training fidelity at $r = 0.963$.
 
  \item \textbf{The emulator is a reliable predictor.} Hardware and
    noisy emulator overlaps correlate at $r = 0.952$ across the 20
    pairs, and the hardware gap ($0.257$) exceeds the emulator
    prediction ($0.238$). This validates the emulator-based analysis
    on the remaining 79~pairs.
 
  \item \textbf{Noise does not destroy the signal.} The noisy emulator
    results ($\mu_{\text{sim}} = 0.358$, $\mu_{\text{dis}} = 0.170$)
    closely match the noiseless results ($0.368$ and $0.174$),
    and the hardware results ($0.414$ and $0.157$) confirm that
    real device noise is manageable.
\end{enumerate}

The overall conclusion is that variational quantum circuits can learn
textual content similarity with results comparable to a BiLSTM. When
executed on the Quantinuum H2-2 trapped-ion processor, the trained
circuits produce state overlaps that reliably separate similar from
dissimilar programme pairs (Cohen's $d = 5.11$), confirming hardware
feasibility. These results suggest that recommender systems provide a
potential application domain for studying parameter-efficient quantum
machine learning models in the NISQ era.

\section{Data and Code Availability}

The code for the experiments presented in this work is publicly available at:
https://github.com/Mina-Abbaszade/Quantum-Text-Similarity. 
The repository includes the full implementation of the quantum similarity models, training procedures, and the text-to-circuit pipeline based on DisCoCirc.
The dataset used in this work consists of publicly available TV programme synopses together with derived pairwise similarity labels. The processed datasets (train, validation, and test splits) are provided in the repository to support reproducibility.
Due to storage considerations, precomputed quantum circuits are not included and can be regenerated using the provided code.
The downstream recommender system used for evaluation is based on proprietary BBC infrastructure and includes user data; therefore, it is not publicly available. This repository focuses solely on the similarity learning component. 

\begin{acks}
The authors would like to thank Stephen Clark, Dimitri Kartsaklis, Tiffany Duneau, and Kin Ian Lo for helpful discussions and support during the development of this work. We also gratefully acknowledge Quantinuum for providing access to the H2 emulator and quantum hardware used in this study.
Mina Abbaszadeh gratefully acknowledges University College London (UCL) for supporting her postdoctoral research position.
This work was supported in part by the Royal Academy of Engineering Research Chair in “Engineered Mathematics for Modelling Typed Structure” (RCSRF2122-14-152), sponsored by University College London (UCL), the British Broadcasting Corporation (BBC), and Quantinuum.
\end{acks}

\bibliographystyle{ACM-Reference-Format}
\bibliography{refs}

\end{document}